\def\arxivversion{1}
\AddToHook{package/iclr2027_conference/before}{%
  \let\arxivOriginalAddcontentsline\addcontentsline
}
\AddToHook{package/iclr2027_conference/after}{%
  \iclrfinalcopy
  \let\addcontentsline\arxivOriginalAddcontentsline
}
\AddToHook{begindocument/before}{%
\raggedbottom

\makeatletter
\renewcommand{\@maketitle}{%
  \begingroup
    \parindent=0pt
    \raggedright
    {\LARGE\bfseries\def\\{\unskip\space\ignorespaces}\@title\par}%
    \vspace{0.8em}%
    {\normalsize\bfseries\@author\par}%
  \endgroup
  \vspace{1em}%
}
\makeatother

\author{%
  \parbox[t]{\textwidth}{%
    \raggedright
    Siyuan~Chen\textsuperscript{1}\quad
    Cai~Zhou\textsuperscript{2}\quad
    Jinrui~Zhang\textsuperscript{3}\quad
    Zhaokang~Liang\textsuperscript{4}\\[0.2em]
    Taku~Komura\textsuperscript{5}\quad
    Wojciech~Matusik\textsuperscript{2}\quad
    Stephen~Bates\textsuperscript{2}\quad
    Tommi~Jaakkola\textsuperscript{2}\\[0.2em]
    Wengong~Jin\textsuperscript{4}\quad
    Peter~Yichen~Chen\textsuperscript{1}\quad
    Minghao~Guo\textsuperscript{2,}\setcounter{footnote}{1}%
    \thanks{Corresponding author.}\\[0.6em]
    \normalfont\small
    \textsuperscript{1}University of British Columbia\quad
    \textsuperscript{2}Massachusetts Institute of Technology\\[0.2em]
    \textsuperscript{3}Carnegie Mellon University\quad
    \textsuperscript{4}Northeastern University\\[0.2em]
    \textsuperscript{5}The University of Hong Kong
  }%
}

\hypersetup{%
  pdfauthor={Siyuan Chen, Cai Zhou, Jinrui Zhang, Zhaokang Liang, Taku Komura,
    Wojciech Matusik, Stephen Bates, Tommi Jaakkola, Wengong Jin,
    Peter Yichen Chen, Minghao Guo}
}

  \hypersetup{%
    pdftitle={From Surfaces to Volumes: Registered Geometry for Protein Representation Learning},
    bookmarksnumbered=true,
    bookmarksopen=false
  }
}
\AddToHook{cmd/maketitle/after}{\lhead{Preprint}}
\documentclass{article}

\usepackage{iclr2027_conference,times}

\usepackage{amsmath,amsfonts,bm}

\def\eqref#1{equation~\ref{#1}}

\def\1{\bm{1}}

\DeclareMathAlphabet{\mathsfit}{\encodingdefault}{\sfdefault}{m}{sl}
\SetMathAlphabet{\mathsfit}{bold}{\encodingdefault}{\sfdefault}{bx}{n}

\usepackage{amsmath}
\usepackage{amssymb}
\usepackage{booktabs}
\usepackage{tabularx}
\usepackage{graphicx}
\usepackage{xcolor}
\usepackage{float}
\usepackage{wrapfig}
\usepackage{hyperref}
\usepackage{url}
\usepackage{subcaption}
\usepackage{colortbl}
\usepackage{cleveref}
\definecolor{LimeGreen}{RGB}{50,205,50}

\title{From Surfaces to Volumes: \\ Registered Geometry for \\ 
Protein Representation Learning}

\author{Anonymous Authors}

\begin{document}

\maketitle

\begin{abstract}
Existing protein geometry models typically represent molecular surfaces using
local geometric features such as sampled points, normals, and curvature. While
effective for capturing exposed molecular shape, these representations do not
explicitly model the volumetric organization beneath the surface or provide a
consistent coordinate system for residue-wise volumetric structure. We
introduce Protein-TetSphere, a registered residue-wise volumetric
representation for proteins. Each protein chain is tetrahedralized to obtain
local volumetric regions associated with individual residues, which are then
registered to a shared fixed-topology tetrahedral reference and represented in
a common Laplacian basis. This registration establishes consistent volumetric
coordinates across residues, enabling local three-dimensional deformation to be
integrated with surface and chemical information in a multimodal protein
representation. We evaluate Protein-TetSphere on ligand-binding pocket classification, protein--protein
interface prediction, and de novo protein binder design. Across the three tasks, Protein-TetSphere improves ligand-binding pocket balanced
accuracy from $0.795$ to $0.826$, Pinder-Pair/Site AUROC from
$0.914/0.852$ to $0.932/0.866$, and binder-design success from $14.95\%$
to $19.90\%$ on the BoltzGen Challenge Set and from $27.62\%$ to $32.19\%$
at the ProtDBench backbone level. These results show that
registered volumetric geometry provides complementary spatial information
beyond molecular surfaces across protein recognition, interaction, and design.
\end{abstract}

\section{Introduction}
\label{sec:introduction}

\begin{figure}[t]
\centering
\includegraphics[width=\linewidth]{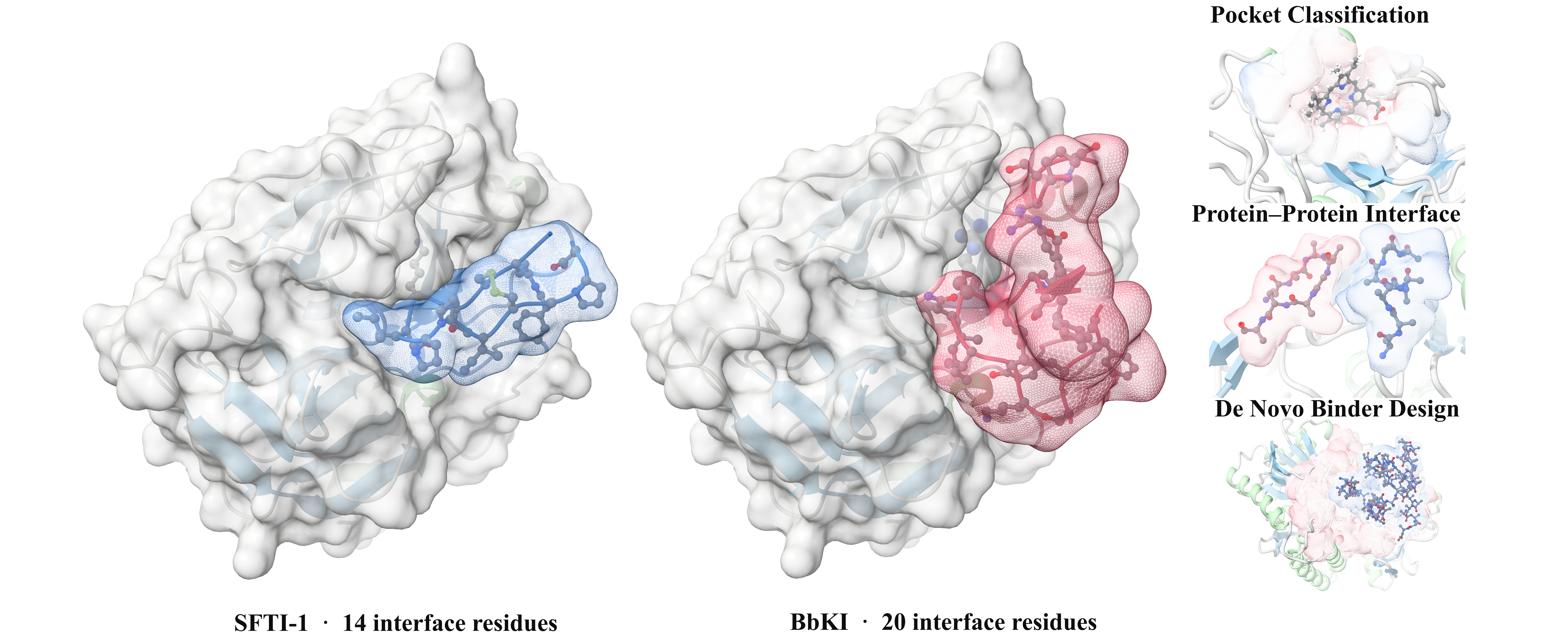}
\caption{\textbf{Motivation of Protein-TetSphere.}
\emph{Left:} SFTI-1 and BbKI illustrate how structurally distinct binders can
engage the same trypsin target through similar target-facing interfaces while
differing substantially in their residue-wise volumetric organization.
\emph{Right:} Protein-TetSphere is evaluated on ligand-binding pocket classification,
protein--protein interface prediction, and de novo protein binder design.}
\label{fig:teaser}
\end{figure}

Protein geometry can be represented at multiple levels. Atomic and residue
graphs organize structure through spatial neighborhoods and directional
relationships~\citep{01_jing2021learning,18_jing2021equivariant}, while surface-based methods represent proteins using points or meshes sampled from the molecular surface together with local geometric features such as normals and curvature~\citep{02_gainza2020deciphering,03_sverrisson2021fast,04_wang2023learning}.
These representations have been effective for capturing exposed molecular shape and have been widely used for recognizing binding pockets and interaction interfaces. Atom-centered geometric models likewise infer interaction sites from local structural neighborhoods~\citep{27_tubiana2022scannet,30_krapp2023pesto}. However, these approaches primarily organize protein geometry through atoms, neighborhoods, or molecular surfaces, and do not explicitly represent how the enclosed three-dimensional volume is distributed among individual residues. Consequently, geometric properties such as residue packing, burial, local
thickness, and the spatial extent of residue-associated regions are only
indirectly captured. Two residues may hence exhibit similar exposed surface
geometry while occupying substantially different three-dimensional volumetric
contexts. As illustrated in Figure~\ref{fig:teaser}, the same trypsin target can
accommodate structurally distinct inhibitors such as SFTI-1 and BbKI through
similar target-facing interfaces despite markedly different underlying
volumetric organization.

Volumetric protein representations based on spatial partitions and voxel grids have demonstrated the utility of internal packing, cavities, and local binding-site structure~\citep{19_rother2009voronoia,20_day2010packing,21_jimenez2017deepsite,29_stepniewska2020improving}.
However, directly extending this idea to residue level introduces a different challenge: residue-associated volumes are irregular and vary substantially in shape and size, so independently constructed local meshes do not provide a common parameterization. A useful residue-wise volumetric representation should therefore preserve local three-dimensional organization while expressing it in a consistent representation space. We hypothesize that registering residue-associated volumes to a shared tetrahedral reference can convert heterogeneous local geometry into a structured family of deformations that can be encoded consistently across a protein.

We introduce \emph{Protein-TetSphere}, a volumetric representation for protein learning that can be registered residue-wise. Given a protein chain, we first construct a tetrahedral representation of the chain and derive residue-specific volumetric regions. Each region is then fitted to a copy of a
shared fixed-topology tetrahedral reference adapted from TetSphere
Splatting~\citep{06_guo2024tetsphere}. Because all fitted regions share the same
connectivity and reference vertex ordering, their deformations can be expressed
in a common Laplacian basis. The resulting coefficients provide a compact and consistent description of
local volumetric geometry that can be processed consistently across residues
and integrated into protein-level models. Protein-TetSphere complements
molecular-surface and chemical features by providing residue-wise
volumetric information in a shared representation.

The same representation supports both discriminative and generative protein learning. We evaluate Protein-TetSphere on ligand-binding pocket classification, protein--protein interface prediction, and de novo protein binder design. For ligand-binding pocket classification, Protein-TetSphere describes the local
volumetric context and packing of residues forming a candidate pocket (\Cref{sec:ligand-results}). For
protein--protein interface prediction, it characterizes residue-level geometry
and packing across interacting chains (\Cref{sec:ppi-results}). Diffusion-based protein design and
inverse-folding models have established complementary routes for generating
backbones and assigning sequences to them
\citep{25_watson2023denovo,26_dauparas2022robust,15_stark2025boltzgen}. For de
novo protein binder design, the registered representation provides geometric
teacher targets that guide the internal representations of a generative model
during training analogously to \cite{14_yu2025representation} (\Cref{sec:design-results}). 
Across these tasks, incorporating the registered volumetric representation improves pocket recognition, interface prediction, and binder-design performance. Together, the results demonstrate that explicitly organizing protein volume in a shared residue-wise coordinate system can provide geometric information complementary to molecular surfaces and offer a general mechanism for incorporating volumetric structure into protein learning.

\section{Related Work}
\label{sec:related-work}

\paragraph{Protein geometric representations.}
Protein structures are commonly represented as atom- or residue-level graphs, where geometric graph neural networks encode spatial neighborhoods, inter-residue distances, and directional relationships through scalar and vector features~\citep{01_jing2021learning,18_jing2021equivariant}. Surface-based
methods instead learn from points or meshes sampled from the molecular surface.
MaSIF learns surface interaction fingerprints~\citep{02_gainza2020deciphering},
dMaSIF provides an efficient end-to-end surface-learning
pipeline~\citep{03_sverrisson2021fast}, and HMR develops harmonic molecular
representations on surface manifolds~\citep{04_wang2023learning}. AtomSurf
integrates graph and surface encoders, enabling atomic neighborhoods and surface geometry to exchange information throughout the network~\citep{05_mallet2025atomsurf}. Other atom-centered approaches, such as ScanNet and PeSTo, represent local structural neighborhoods or atomic point clouds for predicting binding sites and molecular interfaces~\citep{27_tubiana2022scannet,30_krapp2023pesto}. Together, these methods capture complementary aspects of protein geometry through atomic neighborhoods and molecular surfaces, but they do not explicitly represent the enclosed three-dimensional volume at the residue level.

\paragraph{Protein volumetric geometry and tetrahedral representations.}
Volumetric descriptions have long been used to analyze protein interiors.
Voronoi partitioning supports measurements of atomic volume, local packing
density, and internal cavities~\citep{19_rother2009voronoia}, while Delaunay
tessellation has been used to identify recurring tetrahedral residue-packing
motifs~\citep{20_day2010packing}. Grid-based learning provides another route:
DeepSite and Kalasanty represent local protein environments as voxelized
three-dimensional fields for binding-site prediction
\citep{21_jimenez2017deepsite,29_stepniewska2020improving}. These methods
demonstrate the value of spatial occupancy and packing information, but they do
not provide a shared residue-wise deformation coordinate system. In geometric
learning, TetSphere Splatting introduces deformable tetrahedral
spheres as Lagrangian volumetric primitives~\citep{06_guo2024tetsphere}, and
TetCNN studies convolution on tetrahedral meshes~\citep{07_farazi2023tetcnn}.
We adapt tetrahedral volumetric representations to proteins by deriving residue-specific regions and registering them to a shared fixed-topology reference. This provides a parameterization for consistent residue-level volumetric encoding in a shared Laplacian basis.

\paragraph{Protein representation pretraining.}
Protein representation pretraining has been explored from both sequence
and structure. Sequence-based models such as UniRep, ESM, and ProtTrans
learn transferable representations from large-scale unlabeled protein
sequences~\citep{33_ethan2019unified, 34_alexander2021biological, 35_elnaggar2021prottrans}. Protein language models further show that sequence
pretraining can capture structural and functional information
~\citep{34_alexander2021biological}. Structure-aware approaches provide a complementary direction:
GearNet studies structural pretraining~\citep{08_zhang2023protein}, SiamDiff introduces
sequence--structure diffusion objectives~\citep{09_zhang2023pretraining}, and EPT explores equivariant
multimodal molecular pretraining~\citep{10_jiao2026equivariant}. Large-scale pretrained sequence
representations have also enabled protein structure prediction, as demonstrated
by ESMFold~\citep{17_lin2023evolutionary}. We similarly pretrain Protein-TetSphere encoders to learn residue-wise volumetric representations, which are reused across downstream prediction and de novo binder design.
\section{Method}
\label{sec:method}

\begin{figure*}[t]
\centering
\includegraphics[width=\linewidth]{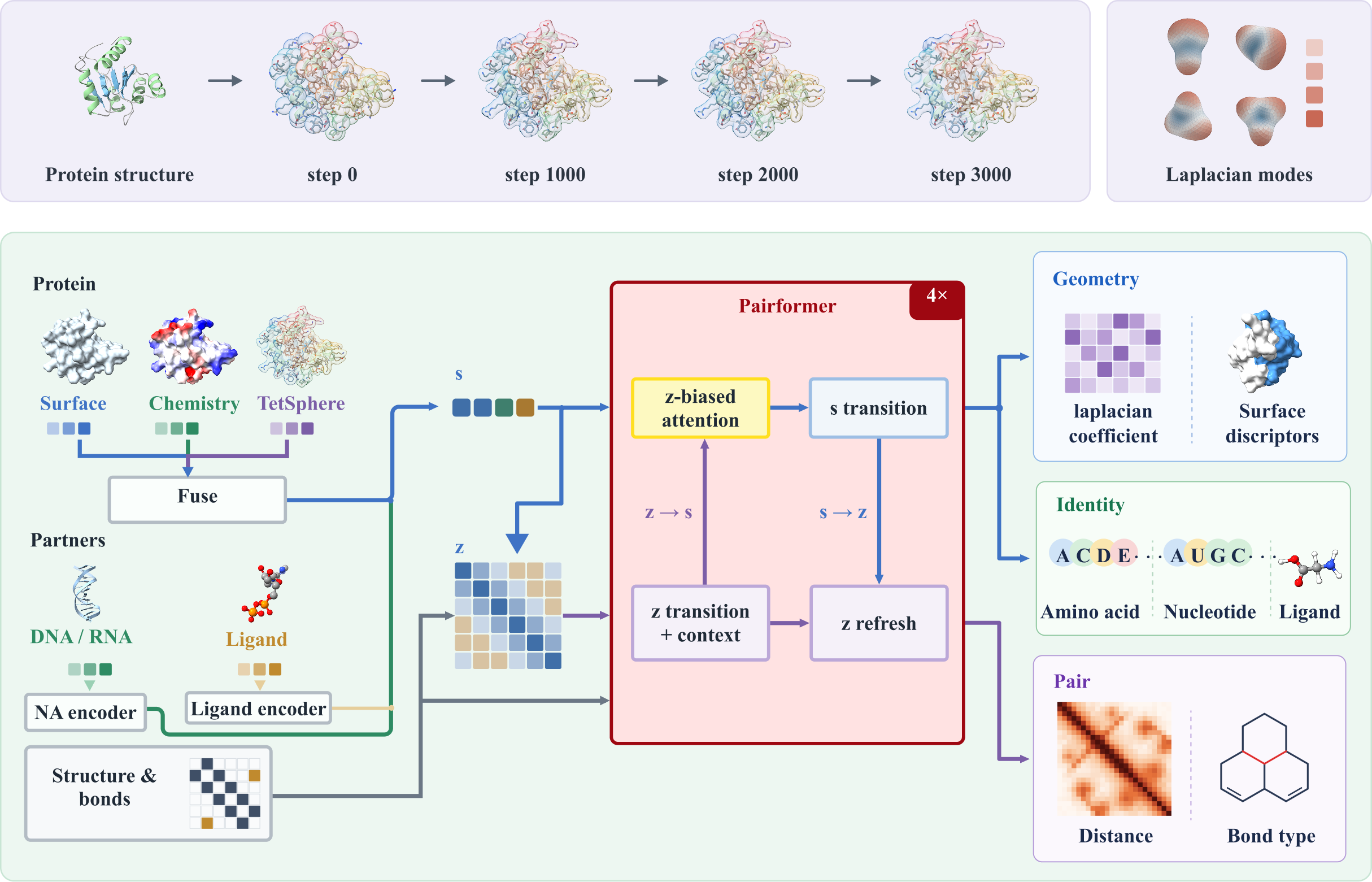}
\caption{\textbf{TetSphere fitting and multimodal pretraining.}
\emph{Top:} per-residue TetSpheres are fitted (steps $0$--$3000$) and
represented by Laplacian coefficients $C_r$.
\emph{Bottom:} surface, chemical, and TetSphere features are fused with
nucleotide, ligand, and bond inputs, processed by a Pairformer-style
single--pair trunk, and pretrained with geometry, identity, and pairwise
structure objectives.}
\label{fig:tetsphere-pipeline}
\end{figure*}

\subsection{Residue-Wise Registered Volumetric Representation}

Given a protein chain and its molecular surface, we construct one fitted tetrahedral mesh for each residue. Let $\mathcal{T}_0=(V_0,\mathcal{K})$ denote a fixed-topology tetrahedral-mesh template sphere~\citep{06_guo2024tetsphere}. For residue $r$ with atom set $\mathcal{A}_r$, we define its geometric center as
\begin{equation}
    c_r =
    \frac{1}{|\mathcal{A}_r|}
    \sum_{a\in\mathcal{A}_r} x_a .
\end{equation}
A copy of the template is placed at $c_r$ and subsequently deformed to match the residue-local molecular geometry; Figure~\ref{fig:tetsphere-pipeline} (top) shows this fitting progression over the course of the optimization.

The fitting proceeds in two stages. We first construct a residue-local molecular surface using Molecular Surface (MSMS)~\citep{36_sanner1996reduced}, and fit the template boundary to this surface using bidirectional point-to-point nearest-neighbor correspondences. The resulting mesh is then refined against the residue region extracted from a tetrahedralization of the complete protein chain, using point-to-triangle closest-point correspondences. In both stages, the template vertices are iteratively updated while preserving the tetrahedral connectivity $\mathcal{K}$, producing a residue-specific fitted tetmesh
\begin{equation}
    \mathcal{T}_r=(V_r,\mathcal{K}).
\end{equation}
Further implementation details are provided in Appendix~\ref{app:implementation}.

\subsection{Spectral Encoding of TetSphere Deformations}

\begin{wrapfigure}{r}{0.48\linewidth}
  \centering
  \vspace{-7pt}
  \includegraphics[width=\linewidth]{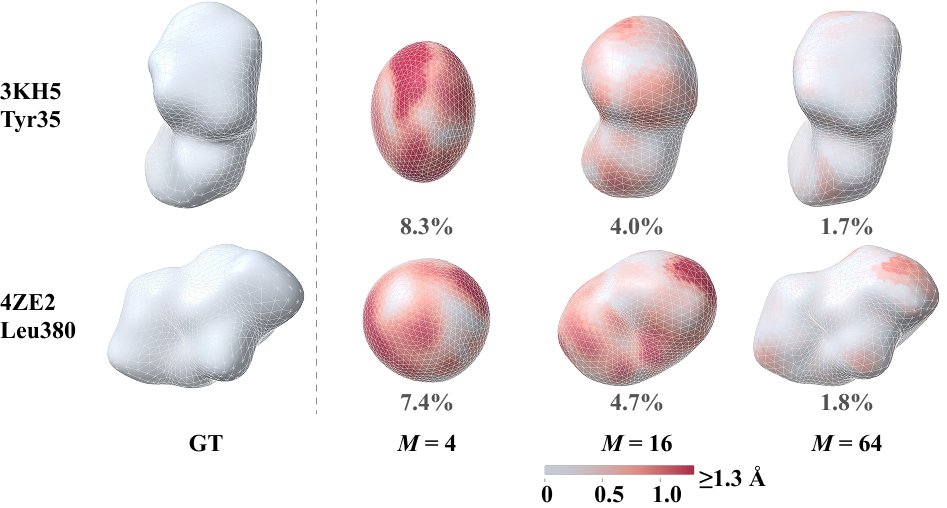}
  \caption{\textbf{Truncated-basis reconstruction.}
Fitted TetSpheres and leading-$M$ Laplacian reconstructions for Tyr35 of 3KH5
and Leu380 of 4ZE2. Colors indicate per-vertex error; values report mean error
relative to residue extent.}
  \label{fig:laplacian-mode-ablation}
  \vspace{4pt}
\end{wrapfigure}

Because all fitted TetSpheres share the same template topology and vertex ordering, their deformation fields can be represented in a common spectral basis. We construct an undirected graph from $\mathcal{K}$ by connecting every pair of vertices that co-occur in a tetrahedron. Let $A$ and $D$ denote the resulting adjacency and degree matrices, respectively. We define the graph Laplacian as
\begin{equation}
    L=D-A,
\end{equation}
and compute its eigendecomposition
\begin{equation}
    LU=U\Lambda.
\end{equation}
We retain the first $M$ non-constant eigenvectors $U_M\in\mathbb{R}^{N\times M}$ to obtain a compact basis for the registered deformation fields; the leading modes are shown in Figure~\ref{fig:tetsphere-pipeline} (top right). 64 modes already reconstruct the fitted volumes to within about two percent of the residue's longest extent (Figure~\ref{fig:laplacian-mode-ablation}).

Let $\Delta V_r\in\mathbb{R}^{N\times3}$ denote the displacement of the fitted TetSphere from its rest-pose template. We project this displacement onto the shared Laplacian basis. To express the resulting deformation vectors in a consistent residue-centered orientation, we construct a right-handed local backbone frame $R_r$ from the N, C$_\alpha$, and C atoms and define
\begin{equation}
    C_r =
    U_M^\top \Delta V_r R_r
    \in\mathbb{R}^{M\times3}.
\end{equation}
Each row of $C_r$ gives the three-dimensional coefficient of a shared deformation mode in the residue-local backbone frame. The matrix compactly describes residue-specific volumetric geometry. Eigenvector sign conventions and reconstruction details are provided in Appendix~\ref{app:implementation}.

\subsection{Protein-TetSphere Representation Pretraining}
We first encode the residue-level spectral coefficients $C_r$ into a TetSphere
embedding. Rather than flattening $C_r$, we treat each spectral mode as a token,
as illustrated in the lower panel of Figure~\ref{fig:tetsphere-pipeline}. For residue $r$ and mode $m$, we construct
\begin{equation}
    h_{r,m}
    =
    \phi_C\!\left(C_r(m,:)\right)
    + e_m
    + \phi_\lambda\!\left(\log(1+\lambda_m)\right),
\end{equation}
where $e_m$ is a learned mode embedding and $\lambda_m$ is the corresponding
Laplacian eigenvalue. A lightweight Transformer processes the mode tokens
$\{h_{r,m}\}_{m=1}^{M}$, which are aggregated by learned-query pooling into a
residue-level TetSphere embedding $t_r$. The TetSphere embedding is then fused
with residue-local surface and chemical representations to form the initial
protein residue features.

To incorporate molecular context, we process these residue features together
with nucleotide and ligand tokens using a shared Pairformer-style single--pair
trunk. Pair features encode structural relations including inter-token distance,
chain and polymer relations, and relative protein geometry. The trunk jointly
updates single representations $s_i$ and pair representations $z_{ij}$, allowing
local surface, chemical, and volumetric information to interact with the
surrounding biomolecular assembly.

We pretrain the model with masked multimodal reconstruction. During training,
selected inputs from the surface, chemical, TetSphere, molecular-identity, and
pairwise streams are withheld, and the model is trained to reconstruct the
corresponding local and relational targets from the remaining modalities and
assembly context. In particular, the TetSphere branch reconstructs the spectral
deformation coefficients, while the other objectives supervise complementary
surface, molecular-identity, distance, and bond information. Further pretraining details are provided in Appendix~\ref{app:multimodal-pretraining}.

\subsection{Adaptation to Protein Learning Tasks}
\label{sec:design-method}

We evaluate a multimodal protein representation that integrates TetSphere
volumetric geometry with surface and chemical information on ligand-pocket
classification, protein--protein interface prediction, and de novo protein
binder design.

For ligand-binding pocket and interface prediction, the protein structure is available
at inference time. For ligand-binding pocket classification, we concatenate the frozen
pretrained single representations $s_i$ with the corresponding AtomSurf residue
features before the original graph-input block, while keeping the remaining
task formulation unchanged. For PPI prediction, we retain the native AtomSurf
site and pair heads and add task-specific residual adapters that predict
site- and pair-level logit corrections from $s_i$, using a symmetric
combination of endpoint representations for residue pairs.

De novo protein binder design has a different information regime because the
geometry of the structure being designed is not available as an input at
inference time. We therefore use REPresentation Alignment
(REPA)~\citep{14_yu2025representation} rather than directly providing
Protein-TetSphere features to the model. REPA introduces an auxiliary training
objective that encourages the internal representations of a generative model
to match features produced by a frozen pretrained teacher, transferring the
teacher's representation without requiring it at inference time.

We apply this strategy to BoltzGen~\citep{15_stark2025boltzgen}, using the
pretrained multimodal model to provide geometric representation targets.
During training, the pretraining model encodes the known protein structures and
provides single and pair targets $(s_i^T,z_{ij}^T)$ for the corresponding
BoltzGen states $(s_i^G,z_{ij}^G)$. Learned projectors map the generator
features into the pretrained representation spaces, and the alignment losses
are added to the native BoltzGen objective:
\begin{equation}
    \mathcal{L}
    =
    \mathcal{L}_{\mathrm{BoltzGen}}
    + \lambda_s \mathcal{L}_{\mathrm{align}}^{\mathrm{single}}
    + \lambda_z \mathcal{L}_{\mathrm{align}}^{\mathrm{pair}}.
\end{equation}
The pretraining model is used only during training and is not required during
design. Task-specific implementation details are provided in
Appendices~\ref{app:ligand-task}--\ref{app:binder-task}.

\section{Experiments}
\label{sec:experiments}

\subsection{Tasks and Evaluation Protocol}
\label{sec:tasks-eval}

We evaluate the multimodal Protein-TetSphere representation on three
tasks: ligand-binding pocket classification, protein--protein interface prediction, and
de novo protein binder design. For ligand-binding pocket classification, we follow the
MaSIF-ligand benchmark introduced by MaSIF~\citep{02_gainza2020deciphering},
use AtomSurf~\citep{05_mallet2025atomsurf} as the surface-aware baseline, and
report balanced accuracy. For protein--protein interface prediction, we follow
the AtomSurf protocol on the clustered PINDER split~\citep{31_kovtun2024pinder}
and evaluate both residue-pair contact prediction (Pinder-Pair) and residue-level
interface prediction (Pinder-Site) using AUROC. For de novo protein binder
design, we transfer the pretrained representation to
BoltzGen~\citep{15_stark2025boltzgen} through representation alignment and
evaluate on the BoltzGen Challenge Set and
ProtDBench~\citep{37_liu2026protdbench}.

We train a separate multimodal pretraining model for each task. For
ligand-binding pocket classification and protein--protein interface prediction, the
model is trained on the corresponding benchmark split, with validation used
for checkpoint selection. For de novo protein binder design, pretraining uses
the released BoltzGen corpus. Detailed task-specific protocols are provided in
Appendices~\ref{app:multimodal-pretraining}--\ref{app:binder-task}.

\subsection{Ligand-binding pocket Classification}
\label{sec:ligand-results}

Ligand-binding pocket classification asks which of seven cofactors occupies a given
protein binding pocket. As shown in Table~\ref{tab:ligand-results}, AtomSurf
achieves a balanced accuracy of $0.795 \pm 0.005$, while the full model improves
this to $0.826 \pm 0.004$. To distinguish the contribution of registered volumetric geometry from
that of an additional trainable representation pathway, we include
\textbf{+ Ours (w/o TetSphere)}. This control retains the trainable Tet token
and the same downstream fusion pathway, but receives no structure-specific
TetSphere input, reaching $0.807 \pm 0.003$. The full model therefore improves
over AtomSurf by $0.031$ and over this control by $0.019$, supporting that the
registered volumetric information contributes beyond the additional latent
capacity alone.

\ifdefined\arxivversion
\begin{table}[htbp]
\centering
\begin{minipage}{0.75\linewidth}
\else
\begin{wraptable}{h}{0.55\linewidth}
\fi
\centering
\setlength{\tabcolsep}{3pt}
\caption{Test balanced accuracy for ligand-binding pocket classification, reported as
the mean and standard deviation over five random seeds.}
\label{tab:ligand-results}
\begin{tabularx}{\linewidth}{@{}>{\raggedright\arraybackslash}Xc@{}}
\toprule
Method & \shortstack{Balanced accuracy $\uparrow$} \\
\midrule
AtomSurf~\citep{05_mallet2025atomsurf} & $0.795 \pm 0.005$ \\
+ Ours (w/o TetSphere) & $0.807 \pm 0.003$ \\
+ Ours & $\mathbf{0.826 \pm 0.004}$ \\
\bottomrule
\end{tabularx}
\ifdefined\arxivversion
\end{minipage}
\end{table}
\else
\end{wraptable}
\fi

\begin{figure}[t]
\centering
\captionsetup[subfigure]{justification=centering,singlelinecheck=false}
\newcommand{\lpcstat}[1]{{\scriptsize\color[HTML]{555555}#1 atoms inside the shells}}

\begin{subfigure}[t]{0.32\linewidth}
  \centering
  \includegraphics[width=\linewidth]{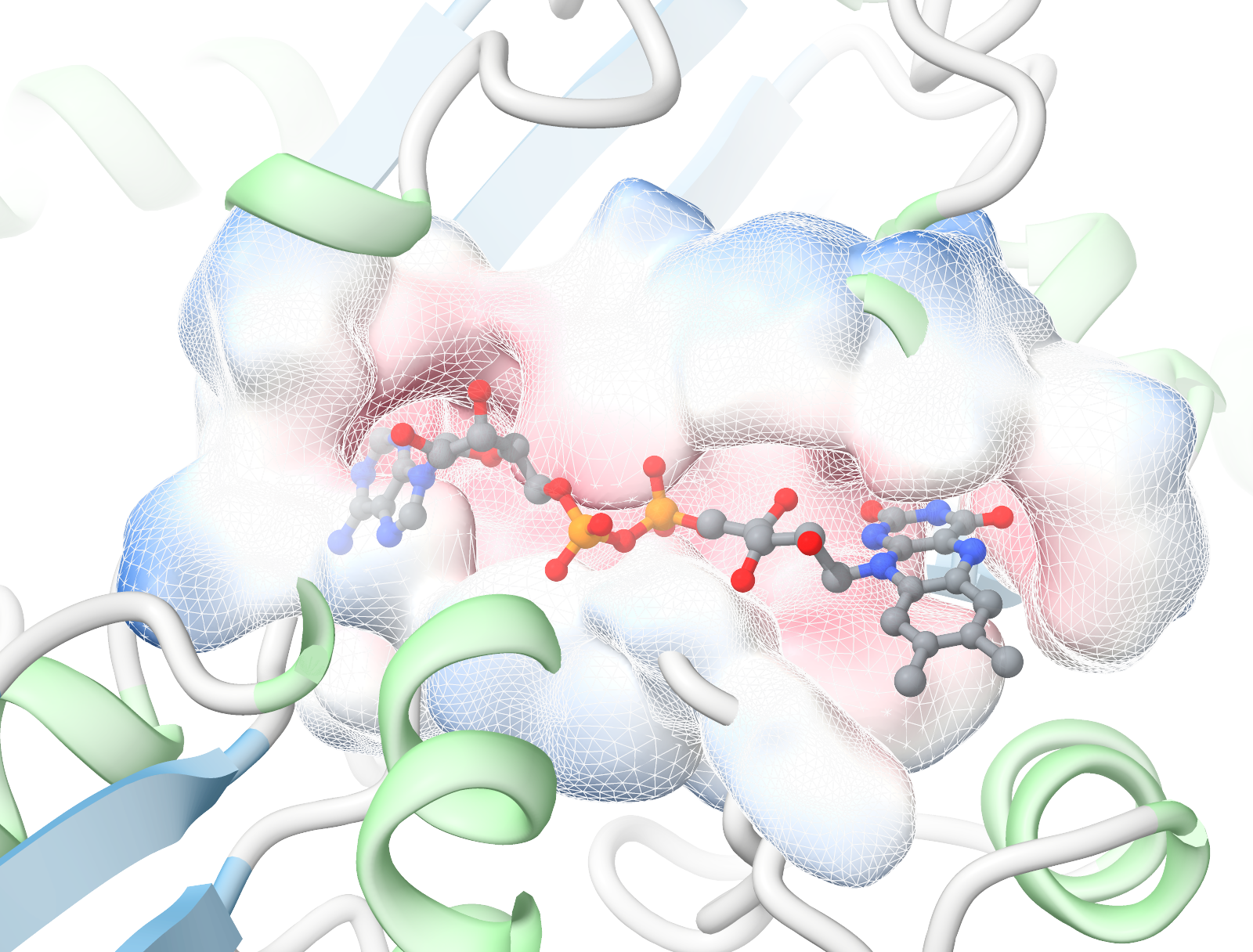}
  \caption{\textbf{\color[HTML]{2E7D4F}FAD} \color[HTML]{2E7D4F}-- native ligand\\
  \lpcstat{0 of 53}}
\end{subfigure}\hfill
\begin{subfigure}[t]{0.32\linewidth}
  \centering
  \includegraphics[width=\linewidth]{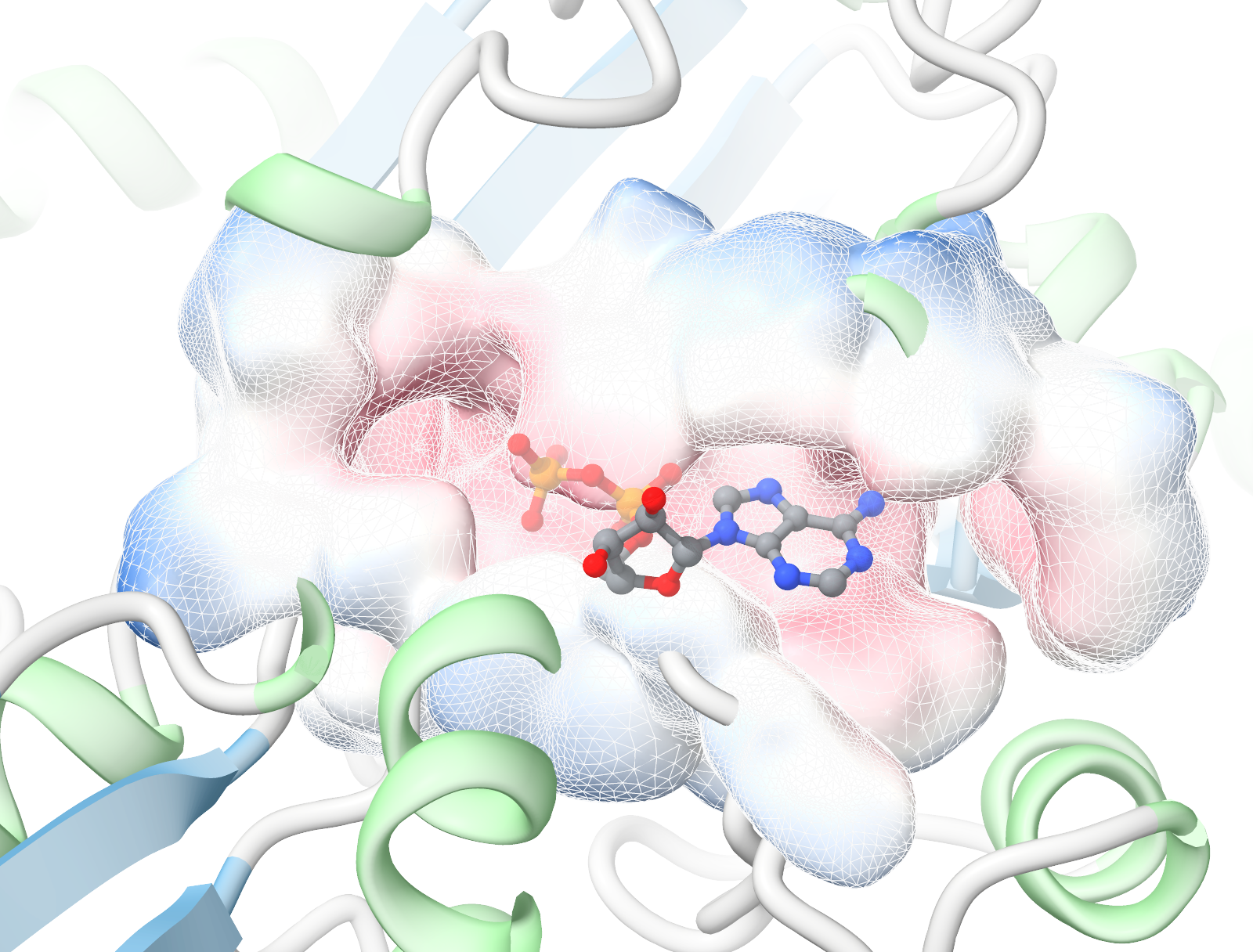}
  \caption{\textbf{\color[HTML]{C76818}ADP} \color[HTML]{C76818}-- too small\\
  \lpcstat{2 of 27}}
\end{subfigure}\hfill
\begin{subfigure}[t]{0.32\linewidth}
  \centering
  \includegraphics[width=\linewidth]{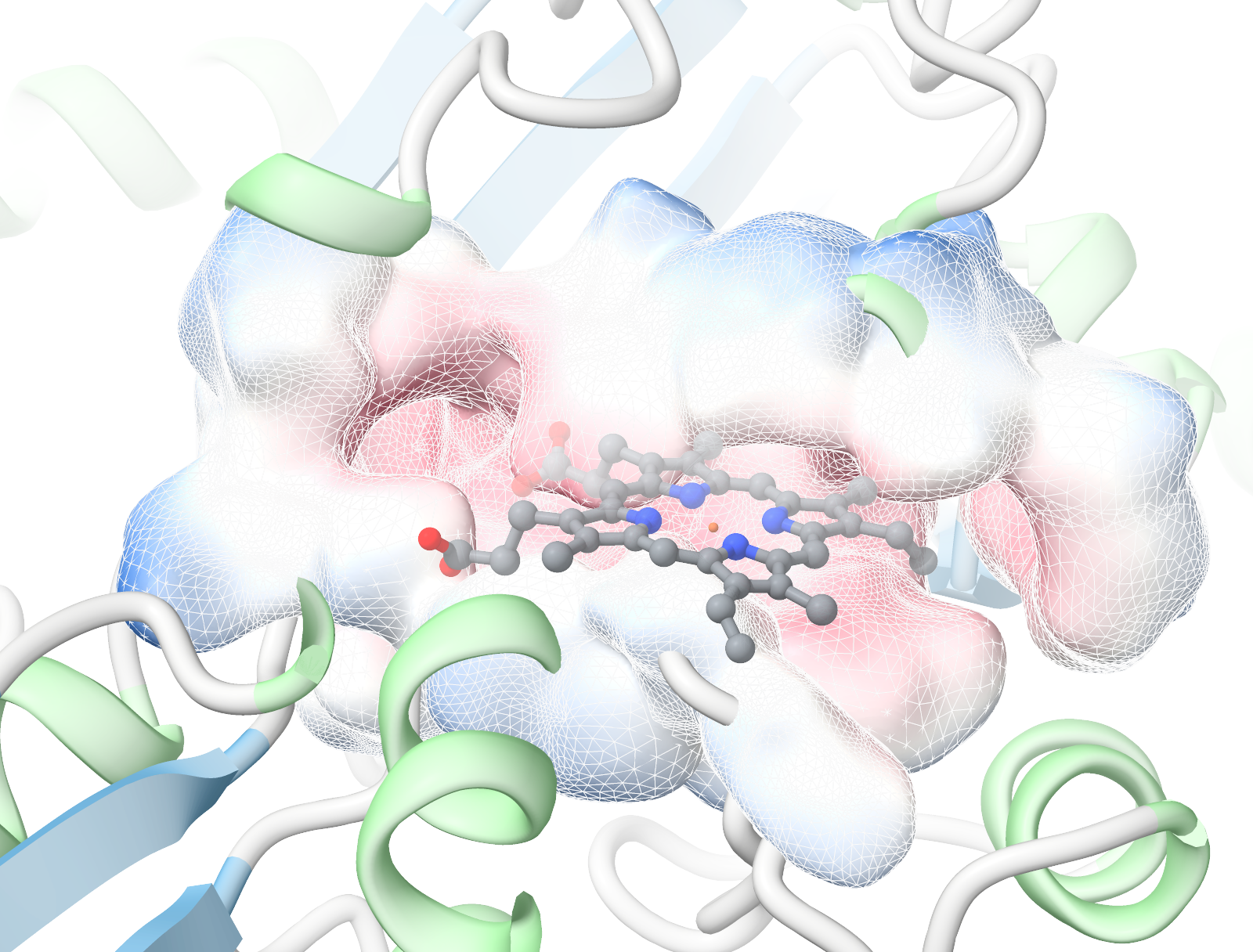}
  \caption{\textbf{\color[HTML]{C76818}HEM} \color[HTML]{C76818}-- intersects\\
  \lpcstat{10 of 43}}
\end{subfigure}

\caption{\textbf{Shape comparison in the 6BD9 FAD pocket.} FAD, ADP, and HEM are compared within the same pocket; numbers report ligand
heavy atoms inside the fitted TetSphere shells.}
\label{fig:ligand-pocket-complementarity}
\end{figure}

Figure~\ref{fig:ligand-pocket-complementarity} illustrates how the registered
volumetric geometry captured by TetSphere can provide complementary cues for
ligand-binding pocket recognition. In this FAD-binding pocket, the native FAD remains
compatible with the fitted residue-wise volumes, whereas ADP under-occupies the
pocket and HEM intersects the fitted shells. Unlike a surface description alone,
TetSphere explicitly represents the three-dimensional organization of the
pocket interior through registered residue-wise volumes, making differences in
volumetric occupancy directly accessible to the model. Consistent with this
geometric distinction, the full model correctly identifies the pocket as
FAD-binding.

\subsection{Protein--Protein Interface Prediction}
\label{sec:ppi-results}

We evaluate protein--protein interface prediction on clustered
PINDER~\citep{31_kovtun2024pinder}, using Pinder-Pair for residue-pair contacts
and Pinder-Site for residue-level interfaces. As shown in
Table~\ref{tab:ppi-results}, AtomSurf achieves AUROC of $0.914 \pm 0.002$ on
Pinder-Pair and $0.852 \pm 0.002$ on Pinder-Site. \textbf{+ Ours (w/o
TetSphere)} uses the same residual-adapter architecture and training protocol
but no structure-specific TetSphere input, reaching $0.919 \pm 0.002$ and
$0.855 \pm 0.001$. The full model improves to $0.932 \pm 0.001$ and
$0.866 \pm 0.001$, gains of $0.018$ and $0.014$ over AtomSurf and $0.013$ and
$0.011$ over the no-TetSphere control. These results support the contribution
of registered volumetric geometry beyond additional latent capacity.

\begin{figure}[ht]
  \centering
  \setlength{\tabcolsep}{2pt}
  \begin{tabular}{@{}ccc@{}}
    \includegraphics[width=0.323\textwidth]{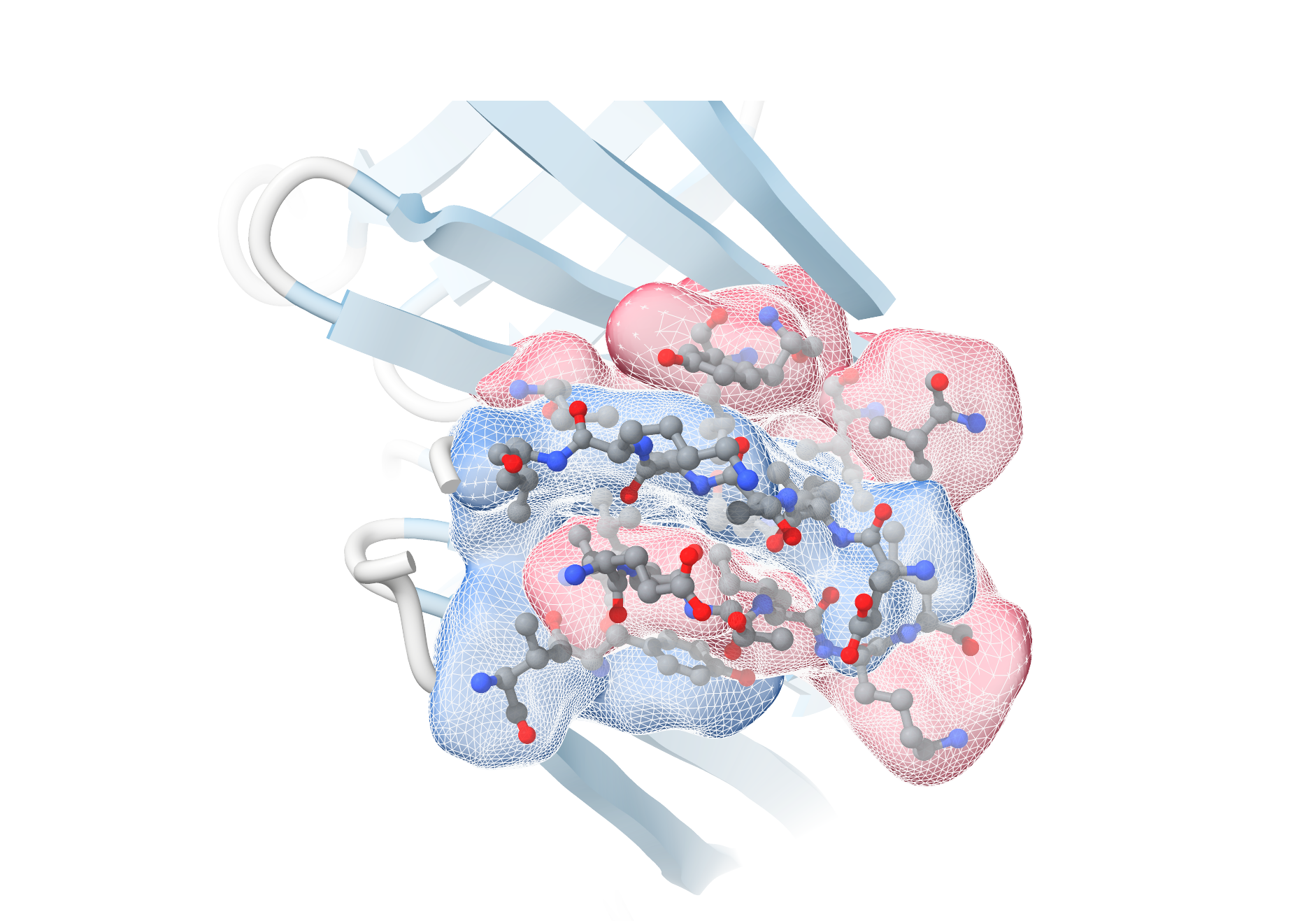} &
    \includegraphics[width=0.323\textwidth]{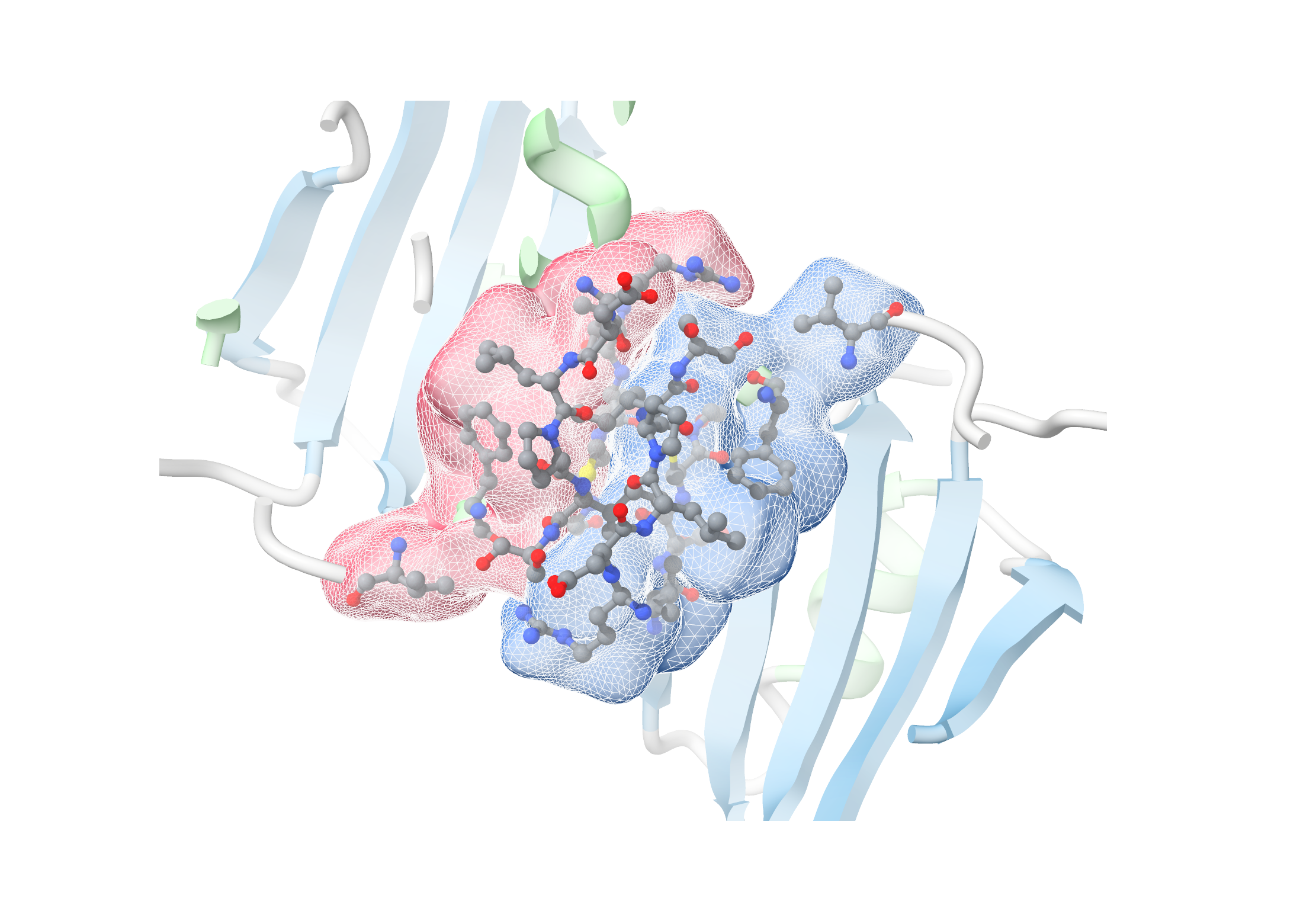} &
    \includegraphics[width=0.323\textwidth]{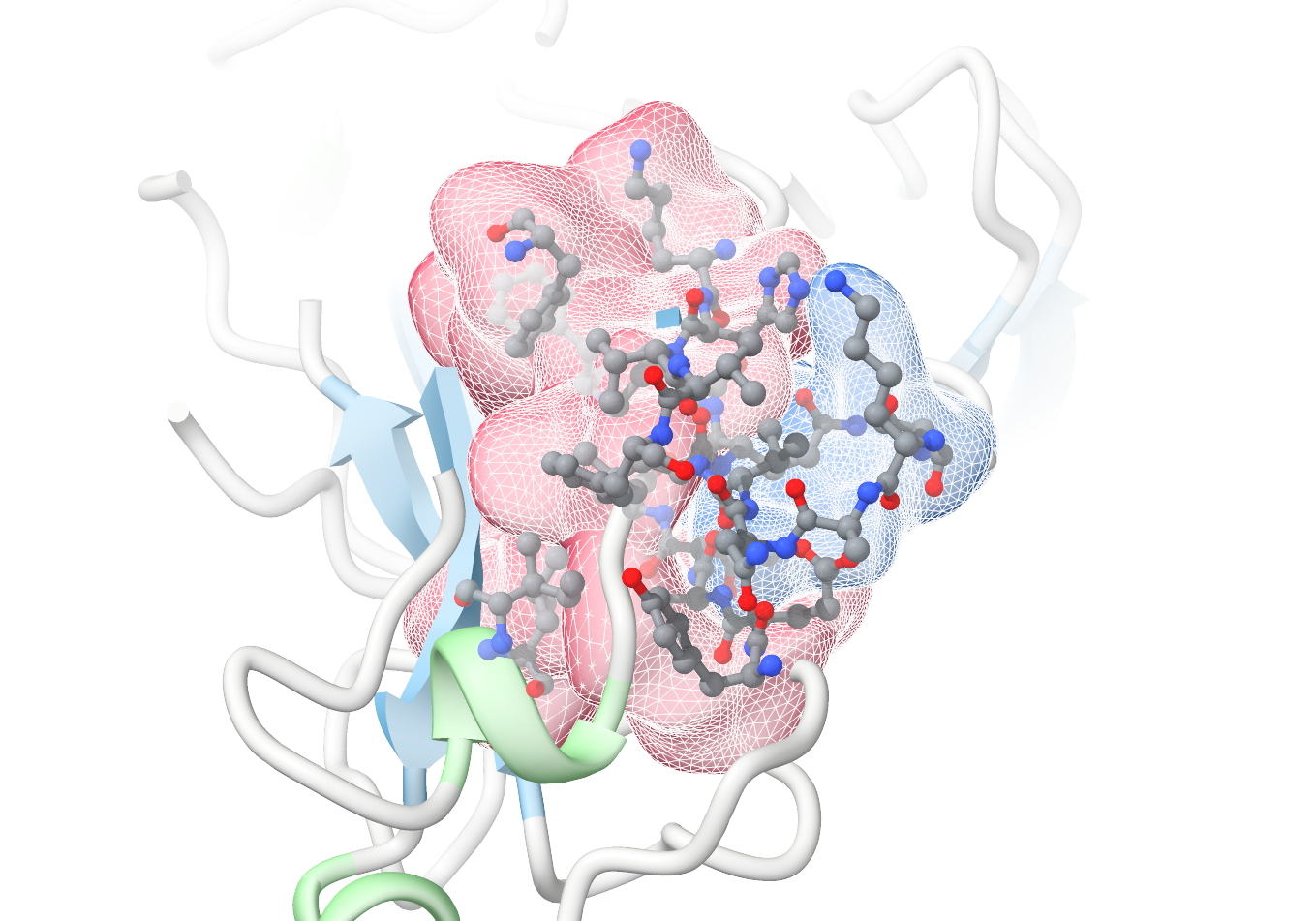} \\
    {\color[HTML]{2E7D4F}\small\bfseries (a) 4MIS} &
    {\color[HTML]{2E7D4F}\small\bfseries (b) 5XOR} &
    {\color[HTML]{2E7D4F}\small\bfseries (c) 1PDK}
  \end{tabular}
  \vspace{-6pt}
\caption{
Representative Pinder-Pair interface predictions from our model. Chain A is
shown in red and chain B in blue, with TetSphere volumes highlighting residues
involved in predicted contacts. Each panel shows 28--30 predicted
contacts, all of which are true contacts under the $5$\,\AA{} heavy-atom
criterion, illustrating the correctness of our high-confidence contact
predictions in these examples.
}
  \label{fig:ppi-interface-patches}
\end{figure}

Unlike residue-site prediction, Pinder-Pair requires identifying specific
residue--residue contacts across the two chains, rather than simply locating
interface residues. Each TetSphere captures residue-local volumetric
deformation in a shared parameterization, and the pretrained representation
integrates these signals with neighboring residues. Together, they provide an
interface view complementary to the surface and chemical features used by
AtomSurf. Figure~\ref{fig:ppi-interface-patches} shows representative
high-confidence Pinder-Pair predictions, where the predicted contacts form
coherent and spatially complementary regions across the interacting chains.

\begin{table}[h]
\centering
\small
\setlength{\tabcolsep}{5pt}
\renewcommand{\arraystretch}{1.05}
\caption{Protein--protein interface prediction on clustered PINDER in the
holo setting. Results are test AUROC (mean $\pm$ s.d.) over five seeds.}
\label{tab:ppi-results}
\begin{tabular}{@{}lcc@{}}
\toprule
Method & Pinder-Pair $\uparrow$ & Pinder-Site $\uparrow$ \\
\midrule
AtomSurf~\citep{05_mallet2025atomsurf}
& $0.914 \pm 0.002$ & $0.852 \pm 0.002$ \\
+ Ours (w/o TetSphere)
& $0.919 \pm 0.002$ & $0.855 \pm 0.001$ \\
+ Ours
& $\mathbf{0.932 \pm 0.001}$ & $\mathbf{0.866 \pm 0.001}$ \\
\bottomrule
\end{tabular}
\end{table}

\subsection{De Novo Protein Binder Design}
\label{sec:design-results}

In de novo binder design, the binder geometry is not available before generation,
so TetSphere cannot be supplied directly as an input. We therefore align
BoltzGen hidden states to the single and pair representations of the multimodal
pretraining model during training, while retaining the native BoltzGen
generation objective. The pretraining model is used only during training.

We evaluate on two ten-target benchmarks, ProtDBench and the BoltzGen Challenge
Set. For ProtDBench, we follow the original evaluation setting, generating 32
backbones for each of 150 (target, binder-length) conditions and eight
ProteinMPNN sequences per backbone, yielding 4,800 backbones and 38,400
sequence candidates per method. We report both backbone-level and
sequence-level pass rates. For the BoltzGen Challenge Set, each method
generates 2,000 candidates using five seeds per target and 40 designs per seed.
All generation budgets and evaluation criteria are held fixed across methods.
Full protocols are provided in Appendix~\ref{app:binder-task}.

\begin{table}[tb]
\centering
\caption{\textbf{Binder-design success across four training arms.}
Rows prefixed with $+$ are cumulative, with a shared denominator within each
block. Denominators are 38{,}400 ProtDBench sequences, 4{,}800 ProtDBench
backbones, and 2{,}000 Challenge Set candidates. Best and second-best values
are shaded dark and light green. The $\pm$ values are bootstrap standard
deviations over designs with targets fixed. See
Appendix~\ref{app:binder-task} for details.}
\label{tab:boltzgen-design}

\small
\setlength{\tabcolsep}{5pt}
\begin{tabularx}{\linewidth}{@{}Xcccc@{}}
\toprule
 & & & +Ours & \\
Criterion or stage & BoltzGen & +Cont. & (w/o Tet) & Ours \\
\midrule

\multicolumn{5}{@{}l}{\textit{ProtDBench: independent pass rate}} \\

Normalized pLDDT
& \cellcolor{LimeGreen!30}94.39\,$\pm$\,0.15
& \cellcolor{LimeGreen}95.18\,$\pm$\,0.17
& 90.49\,$\pm$\,0.22
& 93.97\,$\pm$\,0.19 \\

Interface pTM
& 26.18\,$\pm$\,0.43
& 27.16\,$\pm$\,0.42
& \cellcolor{LimeGreen!30}31.77\,$\pm$\,0.43
& \cellcolor{LimeGreen}32.39\,$\pm$\,0.47 \\

Interface PAE
& 18.23\,$\pm$\,0.38
& 19.60\,$\pm$\,0.38
& \cellcolor{LimeGreen!30}21.42\,$\pm$\,0.38
& \cellcolor{LimeGreen}23.35\,$\pm$\,0.42 \\

\addlinespace
\multicolumn{5}{@{}l}{\textit{ProtDBench: cumulative survival}} \\

All three jointly
& 18.20\,$\pm$\,0.38
& 19.52\,$\pm$\,0.38
& \cellcolor{LimeGreen!30}20.95\,$\pm$\,0.38
& \cellcolor{LimeGreen}23.16\,$\pm$\,0.42 \\

$+$ bound--unbound RMSD (sequence success)
& 13.99\,$\pm$\,0.32
& \cellcolor{LimeGreen}14.80\,$\pm$\,0.33
& 11.73\,$\pm$\,0.31
& \cellcolor{LimeGreen!30}14.67\,$\pm$\,0.34 \\

\addlinespace

Backbone-level success
& 27.62\,$\pm$\,0.50
& \cellcolor{LimeGreen!30}29.29\,$\pm$\,0.52
& 26.46\,$\pm$\,0.55
& \cellcolor{LimeGreen}32.19\,$\pm$\,0.57 \\

\addlinespace
\multicolumn{5}{@{}l}{\textit{BoltzGen Challenge Set: cumulative survival}} \\

Complex backbone RMSD
& \cellcolor{LimeGreen!30}31.35\,$\pm$\,1.03
& 28.55\,$\pm$\,0.95
& \cellcolor{LimeGreen}34.15\,$\pm$\,1.00
& 29.85\,$\pm$\,0.96 \\

$+$ design-subset RMSD
& \cellcolor{LimeGreen!30}30.85\,$\pm$\,1.02
& 27.90\,$\pm$\,0.94
& \cellcolor{LimeGreen}32.65\,$\pm$\,0.98
& 29.10\,$\pm$\,0.95 \\

$+$ binder-only RMSD
& 17.45\,$\pm$\,0.85
& 17.60\,$\pm$\,0.83
& \cellcolor{LimeGreen!30}20.00\,$\pm$\,0.84
& \cellcolor{LimeGreen}23.50\,$\pm$\,0.89 \\

$+$ alanine cap
& 17.00\,$\pm$\,0.84
& 17.60\,$\pm$\,0.83
& \cellcolor{LimeGreen!30}18.90\,$\pm$\,0.82
& \cellcolor{LimeGreen}21.50\,$\pm$\,0.86 \\

$+$ glutamate cap
& 15.05\,$\pm$\,0.79
& 15.45\,$\pm$\,0.79
& \cellcolor{LimeGreen!30}16.95\,$\pm$\,0.79
& \cellcolor{LimeGreen}20.80\,$\pm$\,0.85 \\

$+$ valine cap (final success)
& 14.95\,$\pm$\,0.79
& 15.25\,$\pm$\,0.79
& \cellcolor{LimeGreen!30}16.90\,$\pm$\,0.79
& \cellcolor{LimeGreen}19.90\,$\pm$\,0.84 \\

\addlinespace
\multicolumn{5}{@{}l}{\textit{Diversity-adjusted cluster pass rate}} \\

ProtDBench, TM 0.6
& 8.10
& 10.65
& \cellcolor{LimeGreen}14.00
& \cellcolor{LimeGreen!30}13.65 \\

ProtDBench, TM 0.8
& 16.29
& 20.56
& \cellcolor{LimeGreen!30}22.48
& \cellcolor{LimeGreen}25.10 \\

Challenge Set, TM 0.6
& 8.25
& 7.95
& \cellcolor{LimeGreen}14.70
& \cellcolor{LimeGreen!30}12.25 \\

Challenge Set, TM 0.8
& 11.35
& 10.85
& \cellcolor{LimeGreen!30}15.95
& \cellcolor{LimeGreen}18.40 \\

\bottomrule
\end{tabularx}

\vspace{2pt}
\raggedright\footnotesize
\textit{The decomposition localizes where the endpoint gap emerges; it is not
an additive causal attribution across criteria.}
\end{table}

As shown in Table~\ref{tab:boltzgen-design}, Ours improves the Challenge Set
pass rate from $14.95\%$ for BoltzGen to $19.90\%$, the highest among the four
arms. On ProtDBench, the backbone-level pass rate increases from $27.62\%$ to
$32.19\%$, again the highest result. At the sequence level, Ours reaches $14.67\%$, improving over BoltzGen
($13.99\%$) and remaining comparable to \mbox{+Cont.} ($14.80\%$).

We include two controls to separate volumetric supervision from continued
training and representation alignment. \mbox{+Cont.} continues training
BoltzGen for the same budget without alignment, whereas
\textbf{+ Ours (w/o TetSphere)} uses the same alignment objective and schedule
without structure-specific TetSphere input. The no-TetSphere control reaches
$11.73\%$ sequence-level and $26.46\%$ backbone-level success on ProtDBench
and $16.90\%$ on the Challenge Set. In comparison, the full model reaches
$14.67\%$, $32.19\%$, and $19.90\%$, respectively, showing that registered
volumetric geometry provides additional gains beyond continued training and
representation alignment alone.

\begin{figure}[t]
\centering
\includegraphics[width=0.98\linewidth]{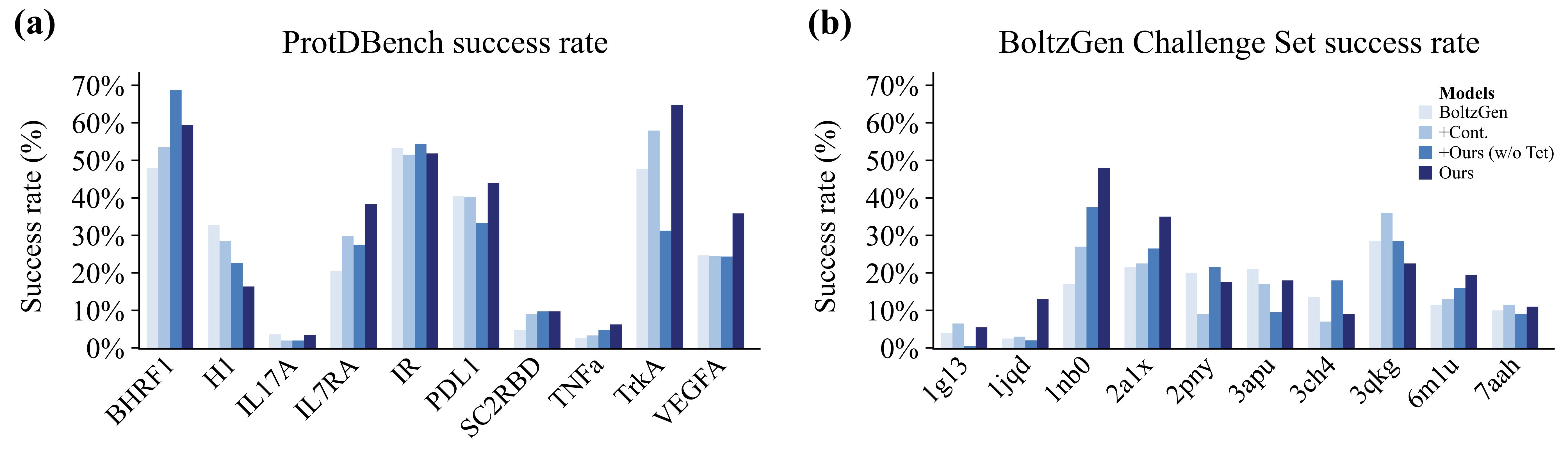}
\caption{Target-level binder-generation success rates, with the same four arms
as Table~\ref{tab:boltzgen-design}: the released BoltzGen checkpoint, the
continued-training control, the no-TetSphere control, and Ours. The left panel
reports ProtDBench at the backbone level, where a backbone counts as a success
if any of its eight redesigned sequences is accepted; the right panel reports
final hard-filter success on the BoltzGen Challenge Set, where each candidate
carries a single sequence. These correspond to the backbone-level success row
and to the final cumulative row of the Challenge Set block in
Table~\ref{tab:boltzgen-design}. Both panels share a common vertical scale so
rates can be read across benchmarks, and both retain all ten targets rather
than collapsing the comparison to one aggregate; response is strongly
target-dependent in both.}
\label{fig:binder-benchmarks}
\end{figure}

The gains are not confined to a small subset of targets. As shown in
Figure~\ref{fig:binder-benchmarks}, Ours outperforms BoltzGen on 7 of 10
ProtDBench targets and 6 of 10 Challenge Set targets, and improves over
\mbox{+Cont.} on 9 and 7 targets, respectively. Relative to the no-TetSphere
control, Ours performs better on 6 of 10 ProtDBench targets and 7 of 10
Challenge Set targets. Thus, the aggregate improvement reflects broadly
distributed gains across both benchmark panels rather than a small number of
outlier targets.

The filter decomposition in Table~\ref{tab:boltzgen-design} further shows where
the gains emerge. On ProtDBench, Ours improves both interface pTM and interface
PAE, increasing the fraction satisfying the first three criteria jointly from
$18.20\%$ to $23.16\%$. The bound--unbound RMSD gate reduces this advantage at
the sequence level, but the gain remains clear at the backbone level, where
success increases from $27.62\%$ to $32.19\%$. This suggests that the improved
interface geometry translates into a larger fraction of backbones with at
least one successful sequence.

On the Challenge Set, Ours shows its largest advantage at the binder-only RMSD
stage: cumulative survival reaches $23.50\%$ compared with $17.45\%$ for
BoltzGen, while the median binder-only RMSD decreases from $1.492$ to
$1.334$\,\AA{}. This advantage is retained through the subsequent composition
filters and leads to the final $19.90\%$ pass rate. Representative successful
designs are shown in Figure~\ref{fig:binder-design-examples} in
Appendix~\ref{app:binder-task}.

The higher success rate is not accompanied by reduced structural diversity.
Ours achieves higher diversity-adjusted cluster pass rates than BoltzGen and
\mbox{+Cont.} on both benchmarks and is highest among all four arms at
TM $0.8$. Although the no-TetSphere control is more diverse at TM $0.6$, its
lower ProtDBench success shows that cluster count alone does not reflect useful
design quality.

Taken together, these results show that volumetric alignment improves
backbone-level success and key geometric quality criteria while preserving
structural diversity, with particularly clear gains in interface quality on
ProtDBench and binder-local geometry on the Challenge Set.

\section{Conclusion and Limitations}
\label{sec:conclusion}

We introduce Protein-TetSphere, a registered residue-wise volumetric
representation of protein geometry. Whole-chain tetrahedralization provides
residue-specific volumetric regions, which are fitted to a shared fixed-topology
reference and encoded in a common Laplacian basis. Ligand-binding pocket and PPI
prediction evaluate these representations through matched AtomSurf comparisons,
while ProtDBench and the BoltzGen Challenge Set evaluate their use as geometric
supervision for binder generation. Across these settings, registered volumetric
geometry consistently complements surface-based representations and improves
protein recognition, interaction prediction, and design.

The current study focuses on static protein structures. A natural extension is to model Protein-TetSphere representations over molecular-dynamics trajectories, capturing time-varying residue-wise volumetric geometry and conformational change. Another direction is to develop multiscale registered representations for biomolecular assemblies, combining residue-level volumes with domain-, complex-, and partner-level geometry. These extensions could broaden volumetric protein representations toward dynamic modeling, interface design, and functional protein generation.

\ifdefined\arxivversion\else
\subsection*{AI use statement}
AI-based tools were used solely to polish sentence-level language, improve
transitions, and standardize the presentation of reported data. They were not
used to generate scientific ideas, design experiments, perform analyses, or make
scientific decisions. The authors reviewed and verified the manuscript and take
full responsibility for its accuracy, integrity, and originality.

\section*{Reproducibility Statement}

We provide descriptions of the model architecture, training procedures, dataset
splits, evaluation protocols, and ablation settings in the main text and
appendices. We will release the code, data splits, and evaluation scripts upon
publication to facilitate reproduction of the results.
\fi

\bibliography{main}
\bibliographystyle{iclr2027_conference}

\newpage
\appendix
\section{Construction and Implementation Details}
\label{app:implementation}
This appendix specifies the deterministic implementation of the Protein-TetSphere construction in Section~\ref{sec:method}.

\subsection{Surface Construction and Coordinate Convention}
For each protein chain, we construct the molecular surface from its present
heavy atoms; no ligand, nucleic acid, solvent, neighboring chain, atom
completion, or OXT synthesis is applied. The MSMS
wrapper uses its standard
PDB-to-xyzrn conversion and invokes MSMS with a $1.5$ Angstrom probe. For a
chain-level MSMS vertex set $S$, the center and normalized coordinate are
\begin{equation}
    c=|S|^{-1}\sum_{x\in S}x,
    \qquad \bar{x}=q(x-c),
    \qquad q=0.04 .
\end{equation}
The chain surface is centered by subtracting $c$ before it is written; target loaders
apply the factor $q$. World-space coordinates are recovered as
$x=\bar{x}/q+c$.

\subsection{MSMS and fTetWild Targets}
For each protein, the surface builder uses its atomic structure to assign a
local atom set to each residue. For a given residue, this set includes the
atoms of the current residue, the previous residue's $C$ and $O$ atoms, and the
next residue's $N$ atoms. MSMS is then run on this set to obtain the
local residue surface. The overlapping atoms only close the local peptide-bond
surface and do not create additional residue rows.

The centered chain-level OBJ is then tetrahedralized with fTetWild.

Let \(Y\) be the TetWild volume vertices and let $A_r$ be the coordinates of the chain atoms belonging
to residue $r$, after subtracting $c$. Each volume vertex receives the label
\begin{equation}
    \ell(y)=\mathop{\arg\min}_{r}\;\min_{a\in A_r}\|y-a\|_2 .
\end{equation}
To make the decomposition cover the interface neighborhood, a radius-$1.0$
ball query on $Y$ propagates every label to nearby volume vertices. For each
residue $r$, the splitter keeps every tetrahedron with at least one vertex in
the expanded label set, compacts its vertices, and extracts its boundary
triangles. The resulting residue targets can overlap, especially near peptide
and chain interfaces, but share the same centered coordinates.

\subsection{Reference Template and Fitting}
Our reference template contains 532 vertices and 1,727 tetrahedra. For each
aligned residue, the current mesh constructor uses a fixed rest scale
$s=0.6$ and initializes one disconnected copy as
$V_{r,0}=sV_0+\bar{x}_r$, without supplying an additional template rotation.
Tetrahedron indices are offset by 532 for each successive copy, yielding
$R\times532$ vertices and $R\times1{,}727$ tetrahedra.

The local-surface and chain-level fitting procedures use the same fixed
topology. In the local MSMS fit, each residue copy is aligned to its local
MSMS target using bidirectional
point-to-point nearest-neighbor displacements. For a predicted boundary vertex
\(v_i\) and its residue target \(S_r\), the forward displacement is
proportional to \(\operatorname{NN}_{S_r}(v_i)-v_i\). Each target point
additionally contributes a backward displacement from its nearest predicted
boundary vertex within the same residue group. The forward and backward fields
are averaged within each direction and combined with equal weights
\(w_f=w_b=0.5\).

In the chain-level fTetWild fit, each predicted vertex is matched directly to
the residue-labeled triangular surface extracted from the complete-chain
fTetWild mesh. A batched bounding-volume hierarchy (BVH) returns the closest
point on the corresponding target triangles. This fit therefore uses a one-sided
point-to-triangle displacement from each predicted vertex to its closest target
surface point; it does not construct a backward target-to-prediction
correspondence.

The fitting displacement \(d_{\mathrm{icp}}\) is combined with
the smooth-barrier displacement \(d_{\mathrm{reg}}\) computed from the rest
pose:

\begin{equation}
    g=-\bigl(d_{\mathrm{icp}}+4\times10^{-5}d_{\mathrm{reg}}\bigr),
    \qquad V\leftarrow V-0.1\,\operatorname{AdamUniform}(g),
\end{equation}
with smooth-energy coefficient $3\times10^{-4}/R$. The two fitting procedures
use budgets of 1,000 and 3,000 vertex updates, respectively. The chain-level
fTetWild fit continues from the in-memory local MSMS result and resets the
AdamUniform state. Both fitting procedures use convergence-based early
termination within these maximum budgets.

After each update, a Gauss--Seidel projection moves vertices to repair inverted
or collapsed tetrahedra using at most 10 projection sweeps. The configured relative floor is $0.1$ times the
absolute signed volume of each rest tetrahedron. Before serialization, a stricter cleanup performs up to eight
rounds of 50 additional projection sweeps.

\subsection{Spectral Basis and Local Frames}
The graph used for the spectral descriptor has one binary edge for every pair
of template vertices that co-occurs in a tetrahedron. We set
$A_{ij}=1$ for these edges, $D_{ii}=\sum_j A_{ij}$, and $L=D-A$. The template
graph is required to have exactly one zero eigenvalue. We retain the first 64
non-constant eigenvectors. For determinism, the sign of each retained
eigenvector $u_m$ is chosen so that its value at
$p_m=\arg\max_j|u_m(j)|$ is positive.

For residue backbone coordinates $n_r$, $a_r$ (C$\alpha$), and $c_r$ (C),
we construct the right-handed frame
\begin{align}
    x_r &= \frac{c_r-a_r}{\lVert c_r-a_r\rVert_2}, \\
    \widetilde{y}_r &= (n_r-a_r)-((n_r-a_r)^\top x_r)x_r, \\
    y_r &= \frac{\widetilde{y}_r}{\lVert\widetilde{y}_r\rVert_2},
    \qquad z_r=x_r\times y_r,
    \qquad R_r=[x_r\;y_r\;z_r].
\end{align}
For each fit,
\begin{equation}
    \Delta V_r=\frac{V_r-sV_0}{q},
    \qquad C_r=U_{64}^{\top}\Delta V_rR_r .
\end{equation}
The omitted constant Laplacian mode corresponds to global translation, which is therefore removed by the projection. Mode/channel means and standard deviations are estimated from valid training
residues only and reused unchanged for validation and test examples. The
inverse low-pass reconstruction is
\begin{equation}
    \widehat{V}_r=sV_0+qU_{64}C_rR_r^{\top}.
\end{equation}

The multimodal pretraining architecture is described in
Appendix~\ref{app:multimodal-pretraining}.

\section{Multimodal Masked-Pretraining Architecture}
\label{app:multimodal-pretraining}

The multimodal pretraining model learns contextual residue representations
within a biomolecular assembly. Protein residues integrate local surface
geometry, amino-acid-derived chemistry, and registered TetSphere deformations,
while nucleotide and ligand tokens provide partner context when present. Each
modality is first encoded according to its own input contract, after which the
resulting tokens interact through shared single and pair states. Throughout
this section, \emph{multimodal pretraining model} refers to this
representation-learning model.

Let $P$, $Q$, and $A$ be the numbers of protein-residue, DNA/RNA nucleotide,
and ligand-atom tokens, respectively, and let $N=P+Q+A$. The encoder returns
contextualized states at both token and token-pair resolution,
\begin{equation}
    Z_{\mathrm{protein}}\in\mathbb{R}^{P\times256},\qquad
    Z_{\mathrm{all}}\in\mathbb{R}^{N\times256},\qquad
    z\in\mathbb{R}^{N\times N\times128}.
\end{equation}
Protein residues are the primary representation targets. Partner tokens supply
assembly context and receive type-specific identity supervision, but are not
assigned synthetic protein-surface, TetSphere, or chemistry inputs.

\subsection{Type-Specific Tokenization}
Table~\ref{tab:pretrain-inputs} summarizes the modality contracts, masks, and
reconstruction targets. Each protein residue is represented by three parallel
views. The Surface
branch encodes 16 residue-local representative points and four normalized
surface descriptors with three invariant message-passing blocks. Its
nearest-neighbor graph is constructed independently within each residue and,
because a residue contributes at most 16 points, is complete: each point
attends to all other points of the same residue. Information therefore cannot
pass between residue surfaces before masking
and fusion. Attention pooling combines the point states with the descriptor
projection to obtain a 256-dimensional Surface token.

The chemistry branch projects hydropathy and charge to the shared width and
contextualizes them with two pair-biased encoder layers. Because these
attributes are derived from amino-acid identity, they are masked jointly with
the corresponding amino-acid target to prevent direct identity leakage. The
TetSphere branch treats the 64 normalized three-channel Laplacian coefficients
as an ordered mode sequence. Each coefficient is projected to a
128-dimensional mode token and augmented with learned mode-index and
eigenvalue embeddings. Two four-head Transformer layers and learned-query
pooling then produce a 256-dimensional TetSphere token. Distinct learned
states represent deliberately masked inputs and genuinely unavailable inputs.
Masking on this branch acts at mode resolution rather than on whole residues:
a residue is first selected with probability $0.15$, and within each selected
residue a further $15\%$ of its valid modes are drawn at random. The drawn
coefficients are zeroed and flagged, while the remaining modes of the same
residue stay visible, so the decoder must exploit correlations across the
spectrum rather than infer a residue shape from nothing. The reconstruction
loss is independent of this draw and is evaluated on every valid mode of every
valid residue.

A learned residue query attends to the Surface, chemistry, and TetSphere
tokens through two residue-local fusion blocks; the updated query becomes the
initial protein state. DNA/RNA nucleotides and ligand atoms use separate
type-specific encoders. Their identity, molecule-type, CCD, and validity
features form 80-dimensional inputs that are independently projected by
two-layer MLPs to the shared 256-dimensional token width.

\begin{table*}[t]
\centering
\caption{Canonical inputs, encoders, masking policies, and reconstruction
targets of the multimodal pretraining model. Task-specific overrides are
described below.}
\label{tab:pretrain-inputs}
\small
\setlength{\tabcolsep}{5pt}
\renewcommand{\arraystretch}{1.12}
\begin{tabularx}{\textwidth}{@{}>{\raggedright\arraybackslash}p{0.13\textwidth}>{\raggedright\arraybackslash}X>{\raggedright\arraybackslash}p{0.15\textwidth}>{\raggedright\arraybackslash}p{0.24\textwidth}@{}}
\toprule
Stream & Input and encoder & Mask policy & Reconstruction target \\
\midrule
Surface
& 16 local points and four descriptors; three residue-local invariant message-passing blocks
& 15\% of valid residues
& Four surface descriptors (Smooth-L1) \\
Chemistry / AA
& Hydropathy and charge; input projection followed by two pair-biased encoder layers
& Joint 20\% AA--chemistry mask
& Amino-acid identity (20-class cross entropy) \\
TetSphere
& 64 three-channel coefficients; mode/eigenvalue embeddings, two Transformer layers, and query pooling
& 15\% of valid residues, then 15\% of the valid modes within each
& Normalized Laplacian coefficients over all valid modes (Smooth-L1) \\
DNA / RNA
& Base, molecule-type, CCD, and validity features; nucleotide-specific two-layer MLP
& 20\% of valid nucleotides
& Base and nucleotide-CCD identities (cross entropy) \\
Ligand
& Element, atom-name, molecule-type, CCD, and validity features; ligand-specific two-layer MLP
& 20\% of valid atoms and components
& Element, atom-name, and component-CCD identities (cross entropy) \\
Relations
& Ordered structural pair features, molecule-type pairs, and ligand-bond inputs
& Sampled valid pairs; supervised bond inputs hidden
& Pair distance (Smooth-L1) and bond type (cross entropy) \\
\bottomrule
\end{tabularx}
\end{table*}

\subsection{Joint Single--Pair Contextualization}
Protein, nucleotide, and ligand states are restored to the native crop order to
form $s^{(0)}\in\mathbb{R}^{N\times256}$. For each ordered token pair, a
42-dimensional structural feature $g_{ij}$ records distance radial-basis
values, signed polymer-relative positions, chain and entity relations,
center validity, the direction from $i$ to $j$ in token $i$'s valid protein
frame, relative protein-frame rotation, and the associated validity flags.
Unavailable coordinates or frames contribute zero-valued geometry together
with explicit validity indicators. Ordered molecule-type and ligand-bond
embeddings provide complementary categorical context.

Pair initialization combines this fixed structural anchor with a
content-derived interaction,
\begin{equation}
    z_{ij}^{\mathrm{mm}}
    =\phi_{\mathrm{pair}}\!\left(
        [s_i+s_j,\,|s_i-s_j|,\,s_i\odot s_j]
      \right),
    \qquad \phi_{\mathrm{pair}}:768\rightarrow256\rightarrow128,
\end{equation}
\begin{equation}
    z_{ij}^{(0)}=\operatorname{LayerNorm}\!\left(
      \phi_{\mathrm{struct}}(g_{ij})
      +e_{\mathrm{type}(i),\mathrm{type}(j)}
      +e_{\mathrm{bond}(i,j)}
      +z_{ij}^{\mathrm{mm}}
    \right).
\end{equation}
Four Pairformer blocks then update the single and pair states jointly. Each
block applies a pair transition, triangle-context aggregation, and a gated
residual from the fixed structural anchor, followed by pair-biased attention
and a transition on the single states and a single-to-pair refresh. Thus,
protein--protein and protein--partner interactions are represented by the same
geometry-aware trunk rather than separate interface modules. Final
normalization produces $Z_{\mathrm{all}}$, and selecting its protein rows gives
$Z_{\mathrm{protein}}$ for protein-specific downstream use.

\subsection{Masked Reconstruction Objective}
Masking forces each representation to recover withheld information from the
remaining modalities and assembly context. Masks are sampled deterministically
for each training draw and applied before the corresponding local or identity
encoder. Amino-acid identity and chemistry share a 20\% residue mask; the Surface input
uses an independent 15\% residue mask; the TetSphere input uses a 15\% residue
mask followed by a 15\% mask over the valid modes of each selected residue; and
nucleotide,
ligand-atom, and ligand-component identities use 20\% masks. When a ligand
component is selected, its CCD input is hidden on all constituent atoms before
their contextualized states are pooled for component prediction.

The canonical objective contains seven weighted terms:
\begin{align}
    \mathcal{L}_{\mathrm{pretrain}}
    ={}&1.00\,\mathcal{L}_{\mathrm{Tet}}
      +0.20\,\mathcal{L}_{\mathrm{Surface}}
      +0.15\,\mathcal{L}_{\mathrm{distance}}
      +0.25\,\mathcal{L}_{\mathrm{AA}} \notag\\
     &+0.20\,\mathcal{L}_{\mathrm{NA}}
      +0.25\,\mathcal{L}_{\mathrm{ligand\text{-}id}}
      +0.10\,\mathcal{L}_{\mathrm{bond}} .
\end{align}
$\mathcal{L}_{\mathrm{Tet}}$ and
$\mathcal{L}_{\mathrm{Surface}}$ reconstruct Laplacian coefficients and
surface descriptors with Smooth-L1 loss. $\mathcal{L}_{\mathrm{Surface}}$ is
restricted to masked residues, whereas $\mathcal{L}_{\mathrm{Tet}}$ supervises
the complete valid coefficient field rather than only the drawn modes; the
error on the masked subset alone is tracked as a diagnostic. $\mathcal{L}_{\mathrm{distance}}$
regresses $\log(1+d_{ij})$ on sampled valid mixed-token pairs.
$\mathcal{L}_{\mathrm{AA}}$ is 20-class cross entropy;
$\mathcal{L}_{\mathrm{NA}}$ combines base and nucleotide-CCD cross entropies in
a $0.7{:}0.3$ ratio; and $\mathcal{L}_{\mathrm{ligand\text{-}id}}$ combines
element, atom-name, and component-CCD cross entropies in a
$0.5{:}0.25{:}0.25$ ratio. $\mathcal{L}_{\mathrm{bond}}$ is five-class cross
entropy over no-bond, single, double, triple, and aromatic labels. Each term is
normalized by its own globally reduced valid-target count. An absent partner
modality therefore contributes neither numerator nor count and does not alter
the stated loss weights.

This stage is a representation pretrainer, not a coordinate or sequence
generator. It uses no MSA, atom diffusion, explicit interface cross-attention,
autoregressive decoder, or full-coordinate reconstruction objective. The
amino-acid classifier is a masked identity head rather than a sequence decoder,
and the chemistry prediction head is retained for diagnostics but is excluded
from the canonical objective. Absolute coordinates enter only through the
shared-frame pair geometry and pair-distance target; TetSphere remains a
translation-free deformation descriptor expressed in residue-local frames.

\subsection{Task-Specific Input Contracts}
The shared single--pair trunk is adapted to the three task settings
through task-specific input contracts. For any pair used for bond
reconstruction, the input bond type is replaced by a learned mask embedding
before pair initialization, so the bond target cannot be copied from the input.

\paragraph{Ligand-binding pocket classification.}
For MaSIF-ligand pretraining, ligand tokens are retained only as masked
context: element, atom-name, and CCD identity fields are replaced by mask
values, while ligand identity, component, and bond targets are disabled. The
ligand coordinate array is retained only to define the pocket instance.
Downstream prediction uses protein geometry alone; ligand coordinates determine
fixed pocket membership and supervision rather than model features.

\paragraph{Protein--protein interface prediction.}
For PPI, the pair representation keeps its 42-channel width, but 30 geometry
channels are zeroed for cross-chain pairs before any learnable pair module
consumes them. These are the distance RBF channels $[0{:}16]$, local direction
$[28{:}31]$, relative rotation $[31{:}40]$, and direction/rotation validity
flags $[40{:}42]$. The 12 topology channels $[16{:}28]$ remain available. The
same mask is applied to the protein pair path used by the chemistry encoder and
to the full mixed-token pair path; cross-chain attention remains enabled by the
token-valid pair mask.

\paragraph{De Novo Protein Binder Design.}
Binder design retains the native BoltzGen inputs and conditioning pathway,
including the provided target structure and benchmark-specific design
conditions. No ligand-specific identity mask or PPI cross-chain geometry mask
is applied. During fine-tuning, the multimodal pretraining model provides fixed
single and pair representation targets for alignment; it is not required at
generation time.

\suppressfloats[t]

\begin{figure}[t]
    \centering
    \includegraphics[width=\linewidth]{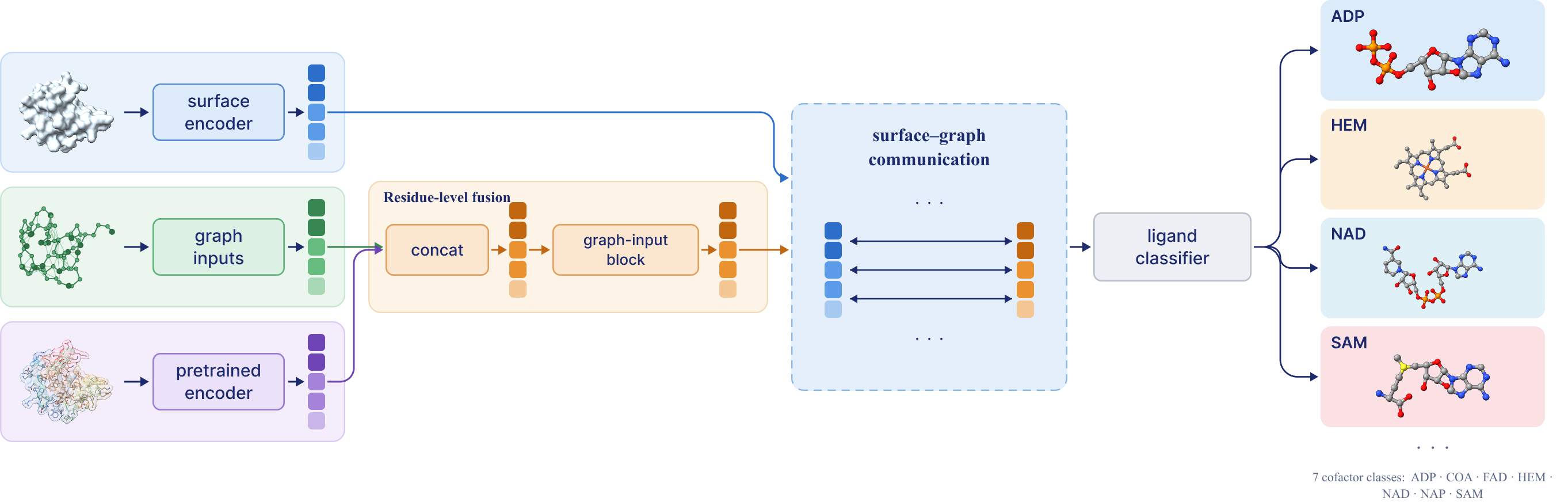}
    \caption{
    \textbf{Ligand-pocket adaptation pipeline.} Adaptation of the pretrained
    residue representation to ligand-binding pocket classification. The native AtomSurf
    molecular surface and residue graph enter unchanged; in parallel, the frozen
    pretrained encoder produces residue-level single representations $s_i$.
    The two are concatenated per residue and consumed by the existing AtomSurf
    graph-input block, so the fused features then follow the original
    surface--graph encoder and ligand-classification head without further
    modification. The same downstream path is used for the full and
    \textbf{+ Ours (w/o TetSphere)} variants; the two differ only in the
    structure-specific input supplied to the frozen pretrained encoder. The
    right-hand column shows four of the seven cofactor classes.
    }
    \label{fig:ligand-fusion}
\end{figure}

\section{Ligand-binding pocket classification}
\label{app:ligand-task}

Given a binding site on a protein, the task is to identify which of seven
cofactors occupies it: ADP, COA, FAD, HEM, NAD, NAP, or SAM. We use the MaSIF-ligand benchmark~\citep{02_gainza2020deciphering}, comprising 1,634 training, 202 validation, and 418 test binding pockets. We retain the standard sequence-based split and report balanced accuracy over the seven ligand classes.

For ligand-binding pocket classification, we train the multimodal pretraining model
using only the training split and select the checkpoint on the validation split. To prevent ligand-identity
leakage during pretraining, the ligand element, atom-name, and CCD-identity
fields are masked, while ligand-identity and component reconstruction targets
and the ligand-bond target are disabled. The only ligand-specific quantity
retained is the coordinate array associated with the pocket instance. For
downstream classification, the model uses protein-only geometry: ligand
coordinates are used only to define the fixed pocket membership and its class
label, consistent with the AtomSurf task formulation. Additional masking details
are given in Appendix~\ref{app:multimodal-pretraining}.

For downstream prediction, we follow the AtomSurf
framework~\citep{05_mallet2025atomsurf}. We implement the baseline using the official public AtomSurf codebase.
Because the exact task architecture used to obtain the results reported in the
original paper is not included in the public release, our absolute baseline
values may differ slightly from the published numbers. We therefore base all
comparisons on the matched implementation used consistently across our
ablation variants. AtomSurf jointly encodes the molecular surface and residue graph. We keep the
pretrained encoder frozen and concatenate its residue-level single
representations with the native AtomSurf residue-graph input features stored in
\texttt{graph.x}, before the AtomSurf encoder.

More precisely, for residue $i$, let
$x_i^{\mathrm{graph}}\in\mathbb{R}^{31}$ denote the native AtomSurf
residue-graph input feature and let $s_i\in\mathbb{R}^{256}$ denote the frozen
pretrained single representation. We form
\[
    \widetilde{x}_i =
    [x_i^{\mathrm{graph}}\,;\,s_i]\in\mathbb{R}^{287},
\]
and use $\widetilde{x}_i$ in place of the original residue-graph input to
AtomSurf's first input block, which projects it to the native
$128$-dimensional graph width. The subsequent surface--graph encoder blocks and
the ligand-classification head are unchanged. The same fusion path is used for
the full and w/o TetSphere variants; only the structure-specific input supplied
to the pretrained encoder differs.

The AtomSurf baseline and our model use the same downstream split, optimization
protocol, checkpoint-selection criterion, and evaluation cohort. Neither model
uses language-model features. For \textbf{+ Ours (w/o TetSphere)}, we retain the same downstream fusion
architecture and replace only the pretrained representation source with the
corresponding No-Tet variant. This control isolates the contribution of the
registered volumetric geometry from that of the additional representation
pathway and model capacity.

For optimization, the pretraining model is initialized from scratch and trained
for 20 epochs with batch size 1 and a maximum of 448 residues per example. We
use AdamW with learning rate $3\times10^{-4}$, $\beta_1=0.9$,
$\beta_2=0.999$, weight decay $10^{-4}$, bf16 autocast, and gradient clipping
at 1.0. The pretraining checkpoint is selected using the validation split only.
The selected model is then used for full-residue inference to export
$256$-dimensional residue-level single representations, which are kept fixed
during downstream training.

All downstream AtomSurf models are initialized from scratch and trained for
200 epochs with a global batch size of 8 using Adam with learning rate
$10^{-3}$, $\beta_1=0.9$, $\beta_2=0.99$, zero weight decay, and gradient
clipping at 1.0. We use a PolynomialLR schedule with 10 warmup epochs and a
minimum learning rate of $10^{-8}$. The ligand-classification objective is
standard seven-class cross entropy. All variants use the same distributed-training configuration. A checkpoint
is saved after every epoch, and the final model is selected by validation
balanced accuracy. The test split is used only for the final evaluation and
does not participate in feature normalization, pretraining checkpoint
selection, or downstream model selection.

\begin{figure}[t]
  \centering
  \setlength{\tabcolsep}{2pt}
  \newcommand{\lcpanel}[3]{%
    \begin{tabular}{@{}c@{}}
      {\color[HTML]{2E7D4F}\small\bfseries #1 -- #2}\\[-1pt]
      {\scriptsize\color[HTML]{555555}ours\,\textcolor[HTML]{2E7D4F}{$\checkmark$}\quad
       AtomSurf: #3\,\textcolor[HTML]{B23A2E}{$\times$}}
    \end{tabular}}
  \begin{tabular}{@{}ccc@{}}
    \includegraphics[width=0.323\textwidth]{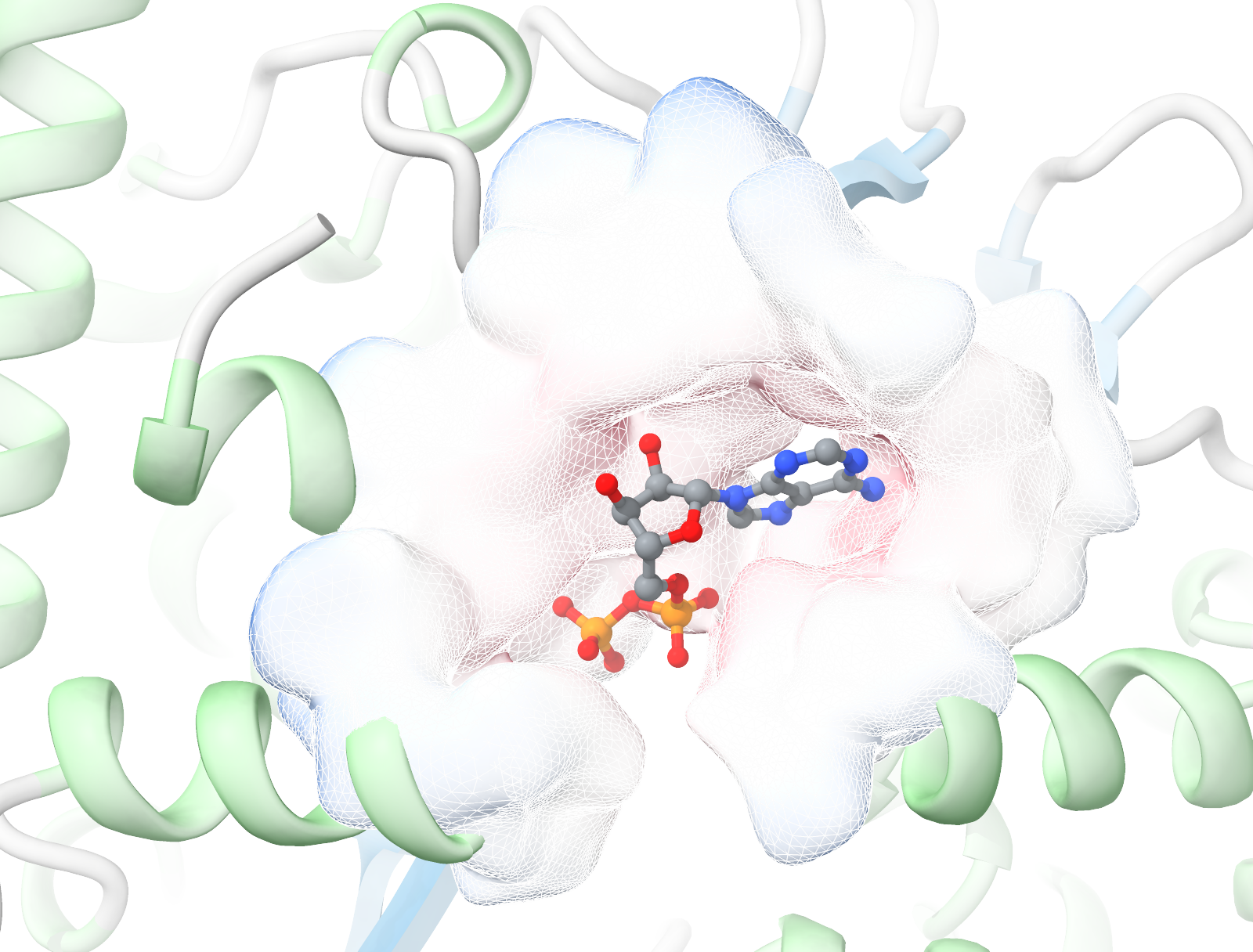} &
    \includegraphics[width=0.323\textwidth]{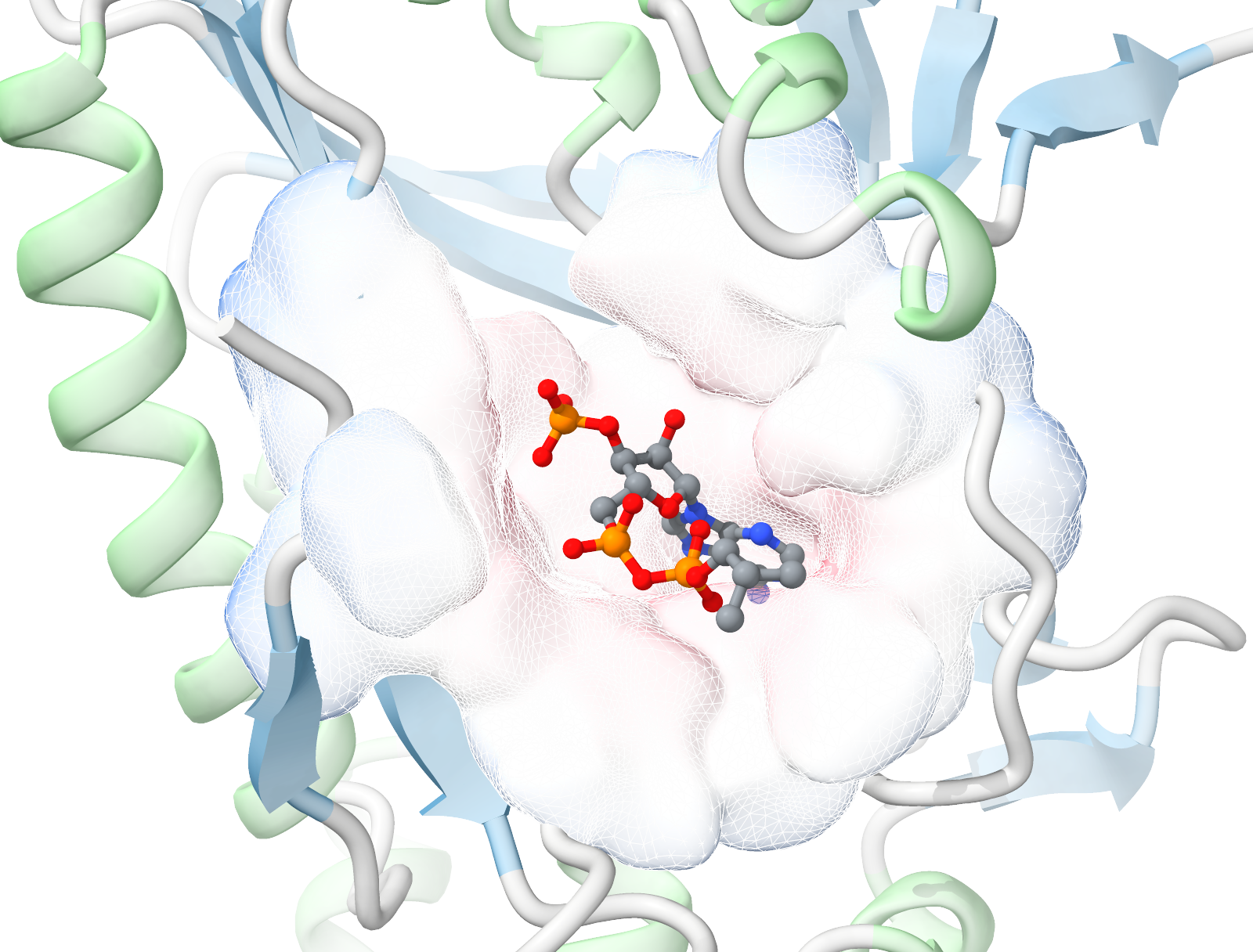} &
    \includegraphics[width=0.323\textwidth]{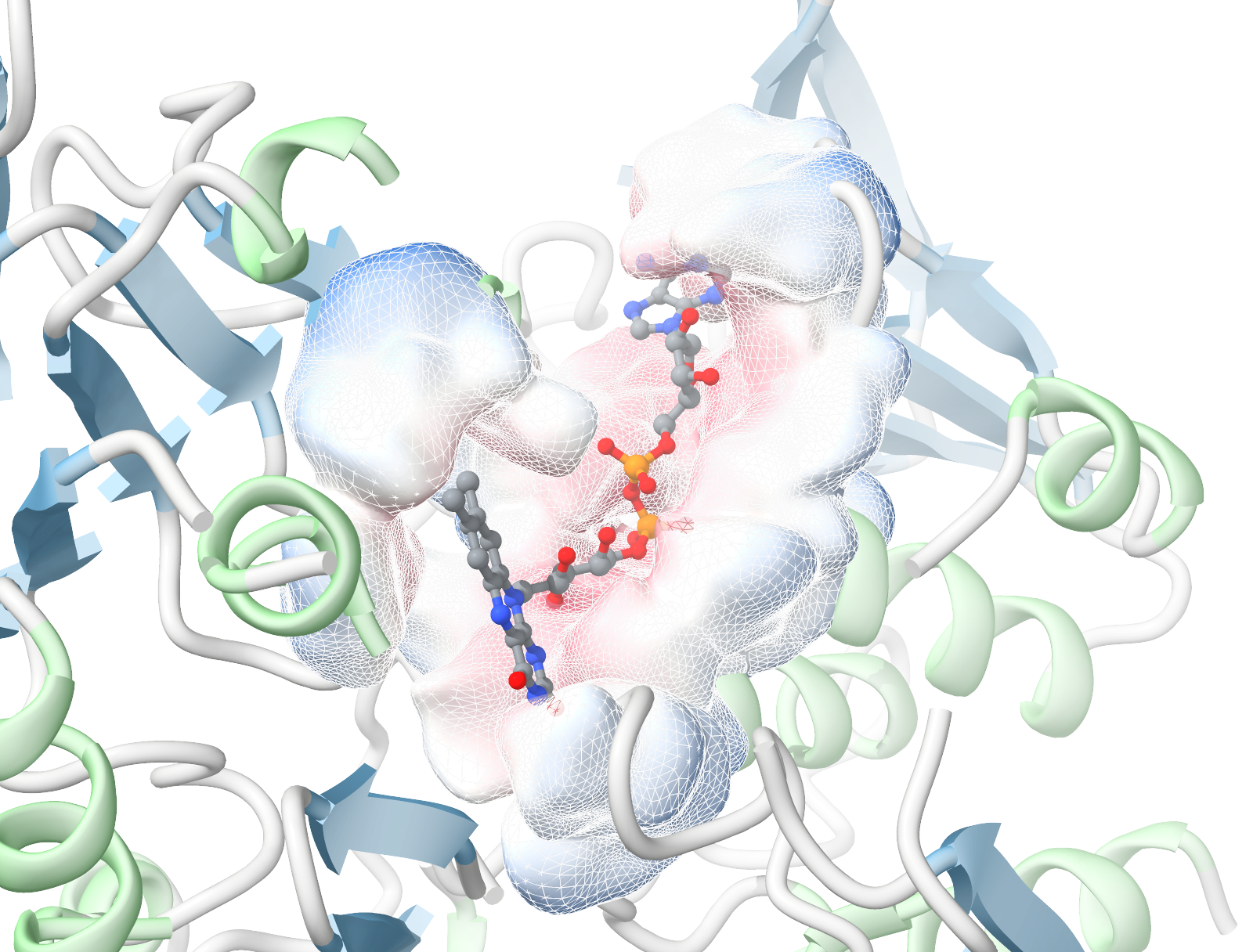} \\
    \lcpanel{(a) 3T4N}{ADP}{SAM} &
    \lcpanel{(b) 3PZC}{COA}{HEM} &
    \lcpanel{(c) 3OC4}{FAD}{SAM} \\[-1pt]
    \includegraphics[width=0.323\textwidth]{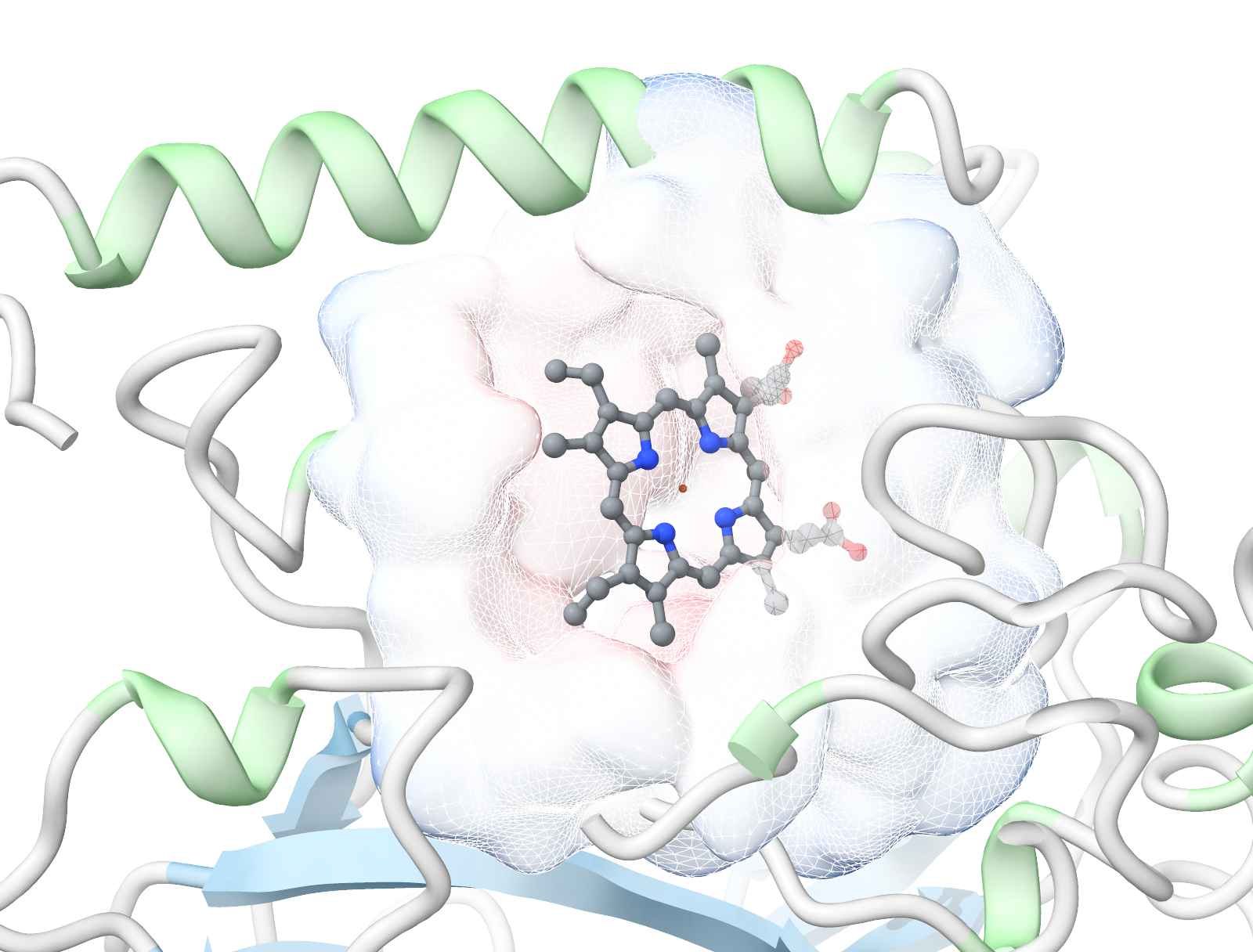} &
    \includegraphics[width=0.323\textwidth]{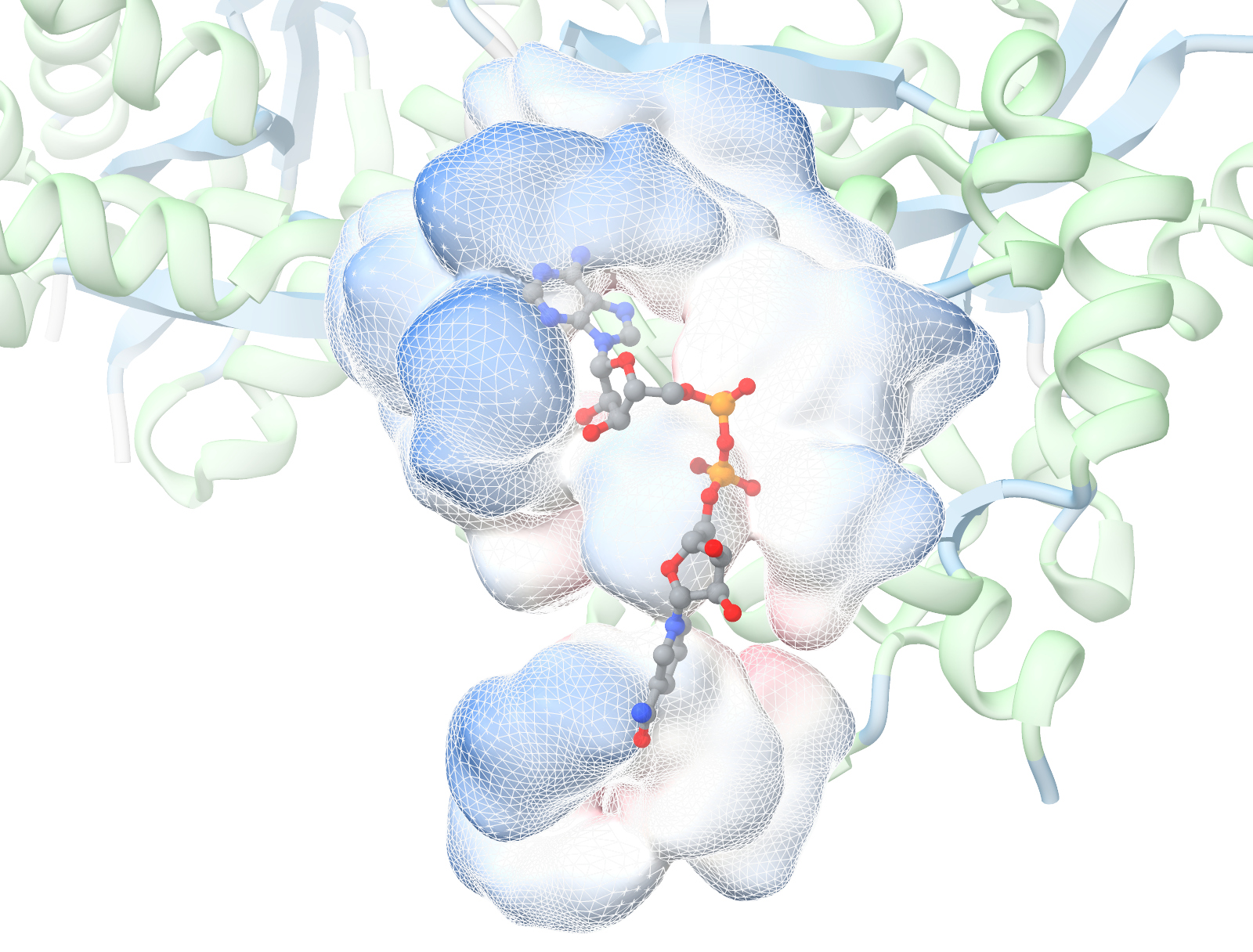} &
    \includegraphics[width=0.323\textwidth]{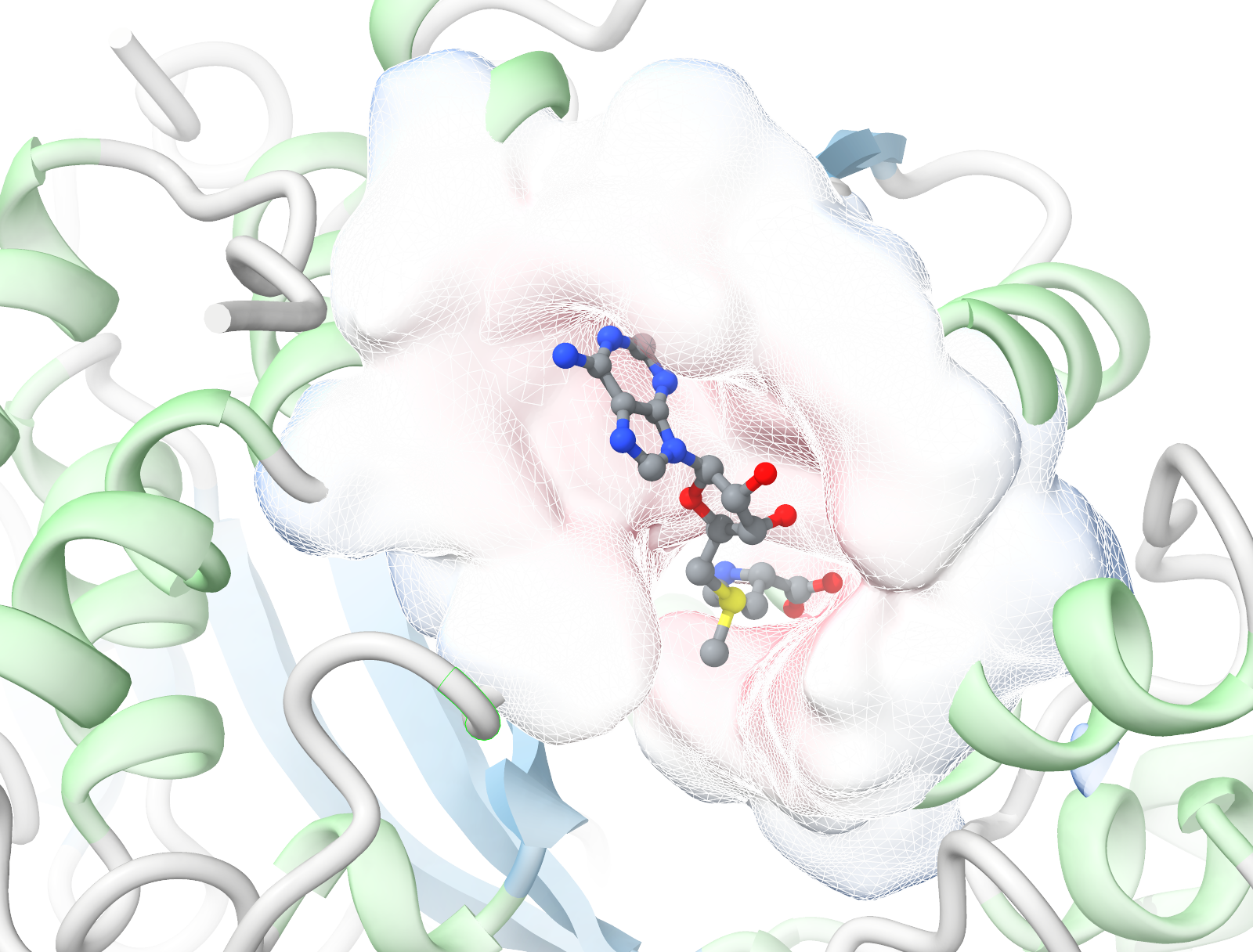} \\
    \lcpanel{(d) 3DY5}{HEM}{FAD} &
    \lcpanel{(e) 1KQN}{NAD}{NAP} &
    \lcpanel{(f) 3AV6}{SAM}{ADP}
  \end{tabular}
  \vspace{-6pt}
  \caption{\textbf{Test pockets our representation classifies correctly and
  native AtomSurf does not.} TetSphere surfaces are colored by distance to the
  native ligand (red close, blue far); labels give the true class and the class
  AtomSurf predicts instead.}
  \label{fig:ligand-classification-examples}
\end{figure}

Figure~\ref{fig:ligand-classification-examples} shows representative test
pockets correctly classified by the full model but misclassified by the
AtomSurf baseline. The examples span multiple ligand classes and are selected
for visual clarity.

\section{Protein--Protein Interface Prediction}
\label{app:ppi-task}

Given two protein chains in their bound conformations, the task is to predict
their interaction interface. Following AtomSurf~\citep{05_mallet2025atomsurf},
we consider two settings. Pinder-Pair predicts whether a residue pair across
the two chains is in contact, whereas Pinder-Site predicts whether an
individual residue belongs to the interface. Both tasks are evaluated using
AUROC over continuous prediction scores.

Our experiments follow the clustered PINDER split, with 42,220 training
clusters and 1,958/1,955 validation/test cluster representatives
~\citep{31_kovtun2024pinder}, and evaluate in the holo setting.

\begin{figure}[t]
    \centering
    \includegraphics[width=\linewidth]{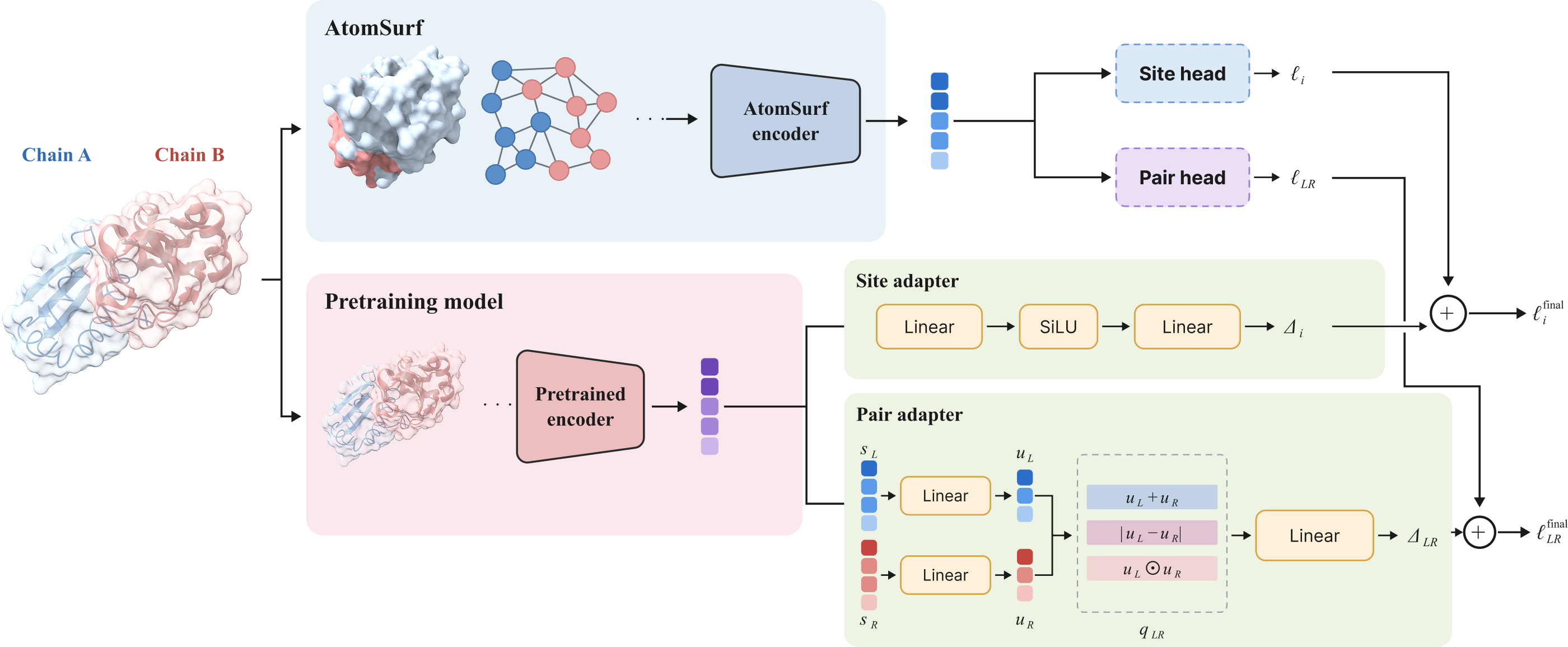}
    \caption{
    \textbf{PPI residual adaptation pipeline.} Residual adaptation of the
    pretrained residue representation for protein--protein interface
    prediction. AtomSurf consumes the molecular surface and residue graph of
    the bound complex and produces the baseline Pinder-Site and Pinder-Pair
    logits $\ell_i$ and $\ell_{LR}$. In parallel, the frozen pretraining model
    yields residue-level single representations, from which a site adapter
    predicts a per-residue correction $\Delta_i$ and a pair adapter predicts a
    residue-pair correction $\Delta_{LR}$ from a symmetric combination of the
    two projected endpoints. Each correction is added to the corresponding
    AtomSurf logit, so the AtomSurf branch is left unchanged. The complex shown
    is barnase--barstar (PDB 1BRS); on the molecular surface, residues within
    $5$~\AA{} of the partner chain are highlighted.
    }
    \label{fig:ppi-fusion}
\end{figure}

The public AtomSurf release does not include the exact task-specific interface
labels used for the originally reported PPI results. We therefore define a residue pair as positive when any heavy-atom distance
between the two residues is below $5$\,\AA{}. This definition is applied
identically to AtomSurf and our model. Because it differs from the private
labeling used for the originally reported AtomSurf results, our absolute scores
should be interpreted within the matched comparison reported here rather than
compared directly with the published AtomSurf numbers.

For each retained system, the positive Pinder-Pair examples are the unique
residue pairs satisfying this contact criterion. We keep all positive pairs and
sample an equal number of non-contact pairs from the complement of the
residue-pair Cartesian product. For Pinder-Site, a residue is labeled as an
interface residue if it appears in at least one positive cross-chain residue
pair. We collect the unique interface residues on each chain and independently
sample an equal number of non-interface residues from the remaining residues on
that chain. The sampling seed is derived deterministically from the split and
system identifier. Training samples are resampled by epoch, whereas validation
and test samples use a fixed sampling state.

The task-specific pretraining model is trained on the PINDER training split,
with validation used for checkpoint selection. Because holo structures directly
expose the relative geometry between the two chains, we mask cross-chain
geometric signals during pretraining to prevent direct leakage of the contact
labels. Specifically, cross-chain distance, local-direction,
relative-rotation, and corresponding validity channels are zeroed before the
learnable pair modules. The remaining topology channels encode non-geometric
relations such as polymer-relative position and chain/entity relationships,
while cross-chain attention remains enabled. Additional details are provided in
Appendix~\ref{app:multimodal-pretraining}.

For downstream prediction, the pretraining model provides one residue-level
single representation
\[
    s_i \in \mathbb{R}^{256}
\]
for each protein residue. Because PPI contains both residue-level and
residue-pair-level prediction tasks, we use task-specific residual adapters
rather than directly fusing the pretrained single representation with the
AtomSurf residue feature, as in ligand-binding pocket classification. This allows the
same pretrained single representations to support residue-wise Pinder-Site
prediction and a symmetric residue-pair construction for Pinder-Pair
prediction. Pretrained pair representations are not passed directly to either
downstream head.

The original AtomSurf Pinder-Site and Pinder-Pair pipelines provide the
baseline logits, while the pretrained single representations are used to
predict additive task-specific corrections.

For Pinder-Site, each residue representation is projected independently:
\[
    u_i^{\mathrm{site}}
    =
    f_{\mathrm{site}}(s_i),
    \qquad
    f_{\mathrm{site}}:
    256 \rightarrow 32,
\]
where $f_{\mathrm{site}}$ consists of LayerNorm, a linear projection, SiLU,
and a second LayerNorm. A residual head maps the resulting 32-dimensional
feature through a $32\rightarrow64\rightarrow1$ multilayer perceptron to
produce
\[
    \Delta_i^{\mathrm{site}}.
\]
The final site logit is
\[
    \ell_i^{\mathrm{site}}
    =
    \ell_i^{\mathrm{AtomSurf}}
    +
    \Delta_i^{\mathrm{site}}.
\]

For Pinder-Pair, the two residue representations are projected using a
separate task-specific adapter:
\[
    u_L^{\mathrm{pair}}
    =
    f_{\mathrm{pair}}(s_L),
    \qquad
    u_R^{\mathrm{pair}}
    =
    f_{\mathrm{pair}}(s_R),
    \qquad
    f_{\mathrm{pair}}:
    256 \rightarrow 32.
\]
We then construct the symmetric interaction feature
\[
    q_{LR}
    =
    \left[
    u_L^{\mathrm{pair}}+u_R^{\mathrm{pair}}\,;\,
    \left|u_L^{\mathrm{pair}}-u_R^{\mathrm{pair}}\right|\,;\,
    u_L^{\mathrm{pair}}\odot u_R^{\mathrm{pair}}
    \right]
    \in\mathbb{R}^{96}.
\]
This construction is invariant to exchanging the two residues. A residual
head maps $q_{LR}$ through a $96\rightarrow64\rightarrow1$ multilayer
perceptron to produce
\[
    \Delta_{LR}^{\mathrm{pair}},
\]
and the final pair logit is
\[
    \ell_{LR}^{\mathrm{pair}}
    =
    \ell_{LR}^{\mathrm{AtomSurf}}
    +
    \Delta_{LR}^{\mathrm{pair}}.
\]

The final linear layers of both residual heads are initialized to zero, so the
residual branches initially contribute zero correction and the model starts
from the original AtomSurf predictions. The full model and
\textbf{+ Ours (w/o TetSphere)} use the same downstream residual architecture;
they differ only in whether structure-specific TetSphere information is
provided to the pretraining model. This control tests whether the improvement
comes from the registered volumetric geometry itself rather than from adding an
additional pretrained representation pathway.

\begin{figure}[t]
  \centering
  \setlength{\tabcolsep}{2pt}
  \begin{tabular}{@{}ccc@{}}
    \includegraphics[width=0.323\textwidth]{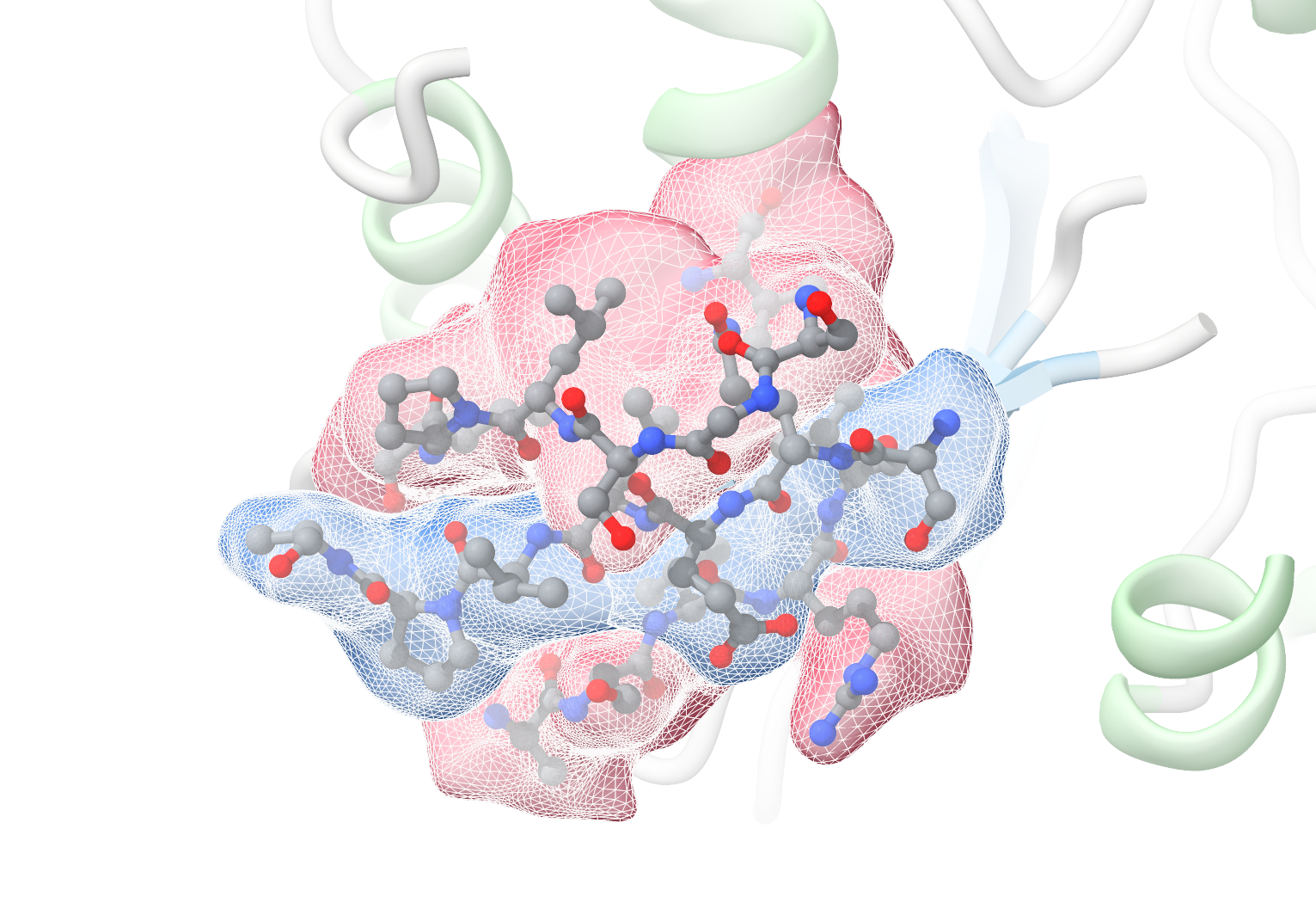} &
    \includegraphics[width=0.323\textwidth]{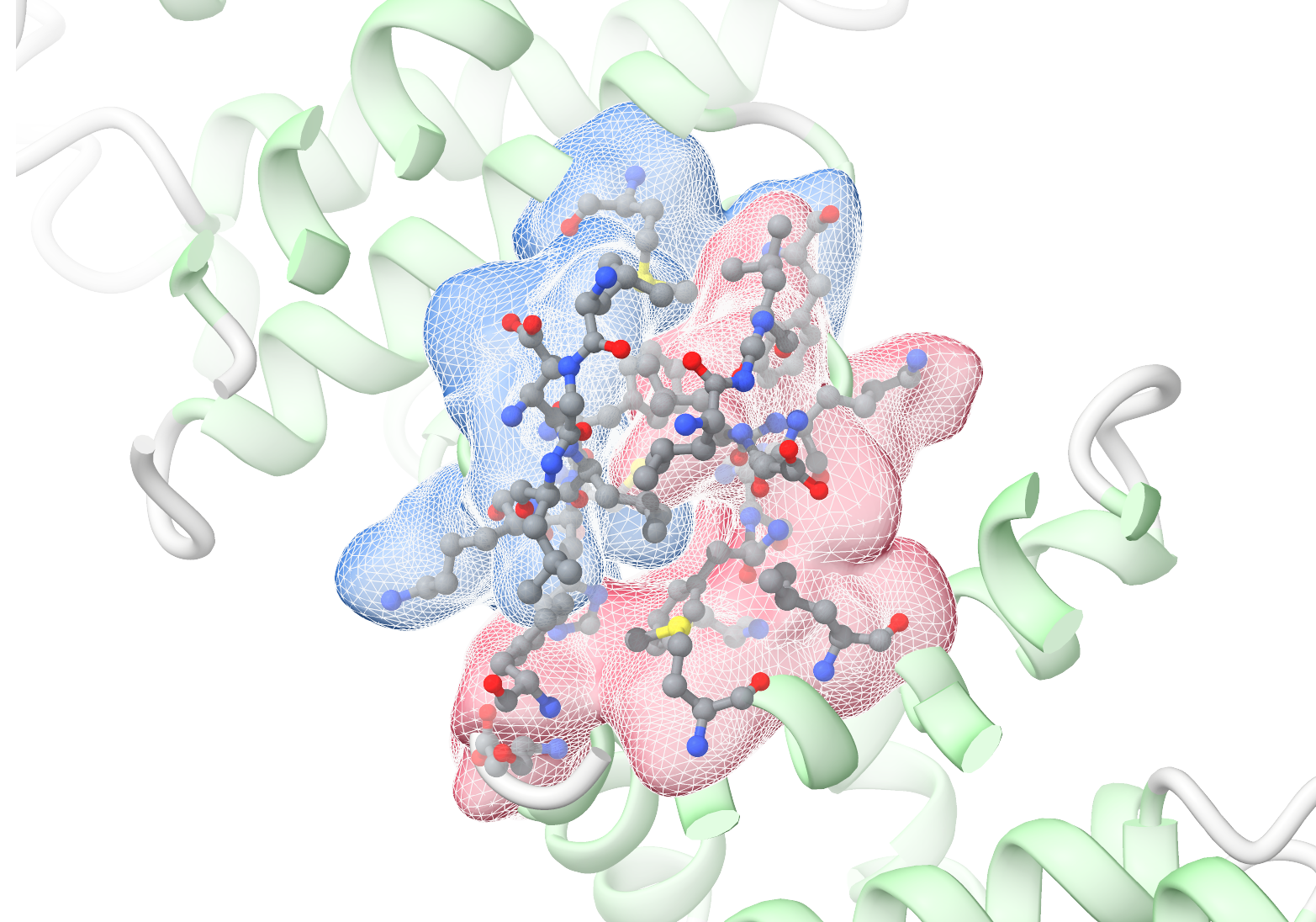} &
    \includegraphics[width=0.323\textwidth]{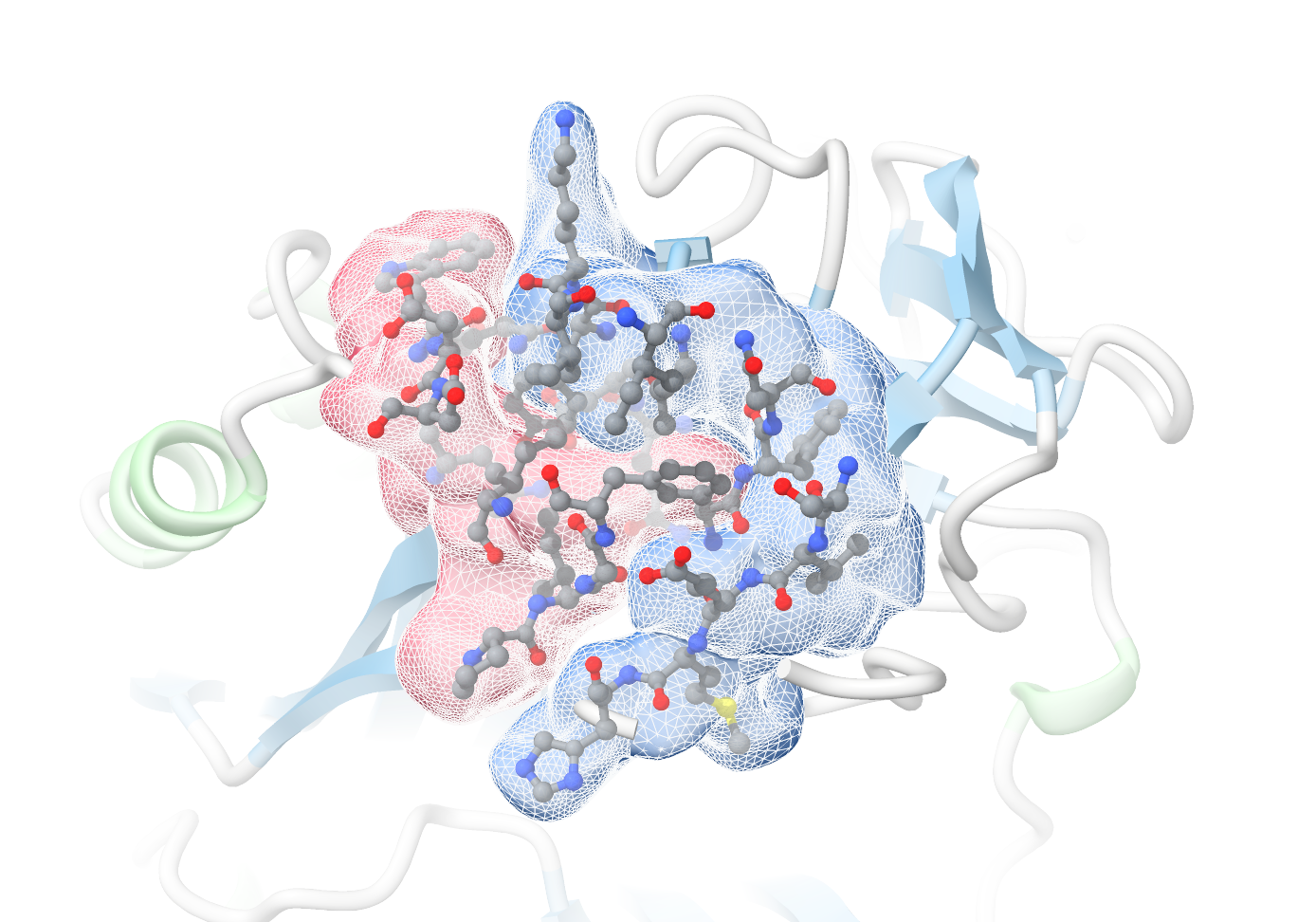} \\
    {\color[HTML]{2E7D4F}\small\bfseries (a) 4HSR} &
    {\color[HTML]{2E7D4F}\small\bfseries (b) 7VPR} &
    {\color[HTML]{2E7D4F}\small\bfseries (c) 1IJE}
  \end{tabular}
  \vspace{-6pt}
  \caption{Further predicted interface regions, drawn exactly as in
  Figure~\ref{fig:ppi-interface-patches}. These three complexes have longer
  chains than the three shown in the main text ($216$--$535$, $280$--$281$ and
  $90$--$438$ residues, against $97$--$215$), so more unrelated structure falls
  in frame behind each patch; the ribbon context is drawn faint and segments
  that would pass in front of the envelopes are removed. Each panel draws $30$
  predicted contacts and all $30$ are true contacts.}
  \label{fig:ppi-interface-patches-appendix}
\end{figure}

For PPI, we train the multimodal pretraining model from random initialization
on the clustered PINDER training split for 60 epochs. We use a batch size of 2
and AdamW with learning rate $10^{-4}$ and weight decay $10^{-2}$, with bf16
precision. Each training example contains at most 512 residues across the two
chains. Complexes within this limit are used in full; for larger complexes, we
apply interface-centered cropping. Residues within $15$\,\AA{} of the partner
chain define candidate interface anchors, from which one anchor on each chain
is sampled and contiguous sequence windows are extracted with a nominal
single-chain neighborhood size of 256 residues. The crop is resampled across
epochs. The PPI-specific masking and reconstruction objectives, including
cross-chain geometry masking and TetSphere partial-mode denoising, follow
Appendix~\ref{app:multimodal-pretraining}. The resulting model is used to
export $256$-dimensional residue-level single representations for downstream
training.

For downstream optimization, all matched PPI experiments use the same fixed
data manifest, sampling protocol, and validation-based model-selection rule.
All downstream variants are trained jointly on Pinder-Site and Pinder-Pair for
up to 100 epochs with batch size 4 across five random seeds. We use Adam with
learning rate $10^{-3}$, $\beta_1=0.9$, $\beta_2=0.99$, zero weight decay, and
gradient clipping at 1.0. The learning rate is controlled by
ReduceLROnPlateau with factor $0.5$, patience 5, and minimum learning rate
$10^{-4}$, using validation Pinder-Site AUROC as the scheduler signal.

The training objective is the equally weighted sum of the Pinder-Site and
Pinder-Pair binary cross-entropy losses. Both terms use BCEWithLogits with a
positive-class weight computed from the globally aggregated positive and
negative element counts,
\[
    w_{+} = \frac{N-N_{+}}{N_{+}},
\]
and are normalized by the corresponding global valid-element counts. Model selection is based exclusively
on validation Pinder-Pair AUROC. 

\section{De Novo Protein Binder Design}
\label{app:binder-task}

De novo protein binder design differs from the two prediction tasks above
because the geometry of the binder being designed is not available before
generation. BoltzGen~\citep{15_stark2025boltzgen} is conditioned on the
provided target structure together with benchmark-defined design conditions,
including target chains, binding-site constraints when applicable, and
binder-length specifications. Consequently, the binder TetSphere
representation, and hence the representation of the completed target--binder
complex, cannot be supplied directly at inference. We therefore transfer the
pretrained geometric representation to BoltzGen through representation
alignment during training.

For this task, we train the multimodal pretraining model described in
Appendix~\ref{app:multimodal-pretraining} on the released BoltzGen PDB data.
Starting from a fixed manifest of 219,627 records, we apply the native BoltzGen
record-level filters: \texttt{SizeFilter} (1--300 chains),
\texttt{DateFilter} (release date before June 1, 2023), and
\texttt{ResolutionFilter} ($0$--$9$\,\AA{}), with predefined validation
records exempted from the training filters. After filtering, 193,326 records
remain in the train/validation split. We further exclude records without
protein chains or without ready protein chains, yielding 189,300 usable
assemblies: 188,902 for training and 398 for validation.

The multimodal pretraining model is trained from random initialization for
160 epochs with batch size 4 using AdamW with learning rate
$3\times10^{-4}$, weight decay $10^{-4}$, and gradient clipping at 1.0.
Training uses bf16 precision. The learning rate uses a 32,000-step warmup,
corresponding to $5\%$ of the configured schedule, followed by cosine decay to
one tenth of the peak learning rate. The checkpoint is selected using the
total validation loss and is kept fixed during downstream BoltzGen training.
Given a training complex with known structure, the resulting model produces
residue-level single and pair representations
$(s_i^{T},z_{ij}^{T})$, which serve as representation targets for the
corresponding internal BoltzGen states.

We initialize the generator from the released BoltzGen checkpoint and align
the final single and ordered-pair representations of the BoltzGen trunk to the
pretrained representation. BoltzGen produces
\[
    s_i^{G}\in\mathbb{R}^{384},
    \qquad
    z_{ij}^{G}\in\mathbb{R}^{128},
\]
while the pretraining model provides
\[
    s_i^{T}\in\mathbb{R}^{256},
    \qquad
    z_{ij}^{T}\in\mathbb{R}^{128}.
\]
The single projector consists of LayerNorm$(384)$, a linear layer from
384 to 384, GELU, and a linear layer from 384 to 256. The pair projector
consists of LayerNorm$(128)$ followed by a linear layer from 128 to 128.

For a projected BoltzGen representation $u$ and its corresponding pretraining
target $v$, both vectors are $\ell_2$-normalized over the feature dimension,
and the alignment error is
\[
    \ell_{\mathrm{align}}(u,v)
    =
    1-\bar{u}^{\mathsf T}\bar{v}
    +
    0.05\,\operatorname{SmoothL1}(\bar{u},\bar{v}).
\]
The single and pair alignment losses are masked means over valid native tokens
and ordered token pairs for which a corresponding pretraining target is
available. Positions without a valid target are excluded from the alignment
term but remain part of the native BoltzGen objective. The complete training
objective is
\[
    \mathcal{L}
    =
    4.0\,\mathcal{L}_{\mathrm{diffusion}}
    +0.05\,\mathcal{L}_{\mathrm{distogram}}
    +0.05\,\mathcal{L}_{\mathrm{align}}^{\mathrm{single}}
    +0.05\,\mathcal{L}_{\mathrm{align}}^{\mathrm{pair}}.
\]

For downstream optimization, we fine-tune the entire BoltzGen trunk together
with the representation projectors. We use AdamW with a peak learning rate of
$2\times10^{-5}$ for the newly initialized projectors and
$2\times10^{-6}$ for the BoltzGen trunk. Both parameter groups use a
1,024-step linear warmup followed by cosine decay to one tenth of their
respective peak learning rates over 20 epochs. Training is performed on eight
H100 GPUs with 8,192 samples per epoch, batch size 1 per rank, diffusion
multiplicity 8, a 512-token crop, and bf16 mixed precision.

The pretraining model is used only to provide representation targets during
training and is not required at inference, leaving the original BoltzGen
generation inputs and native inference procedure unchanged.

\begin{figure}[t]
    \centering
    \includegraphics[width=\linewidth]{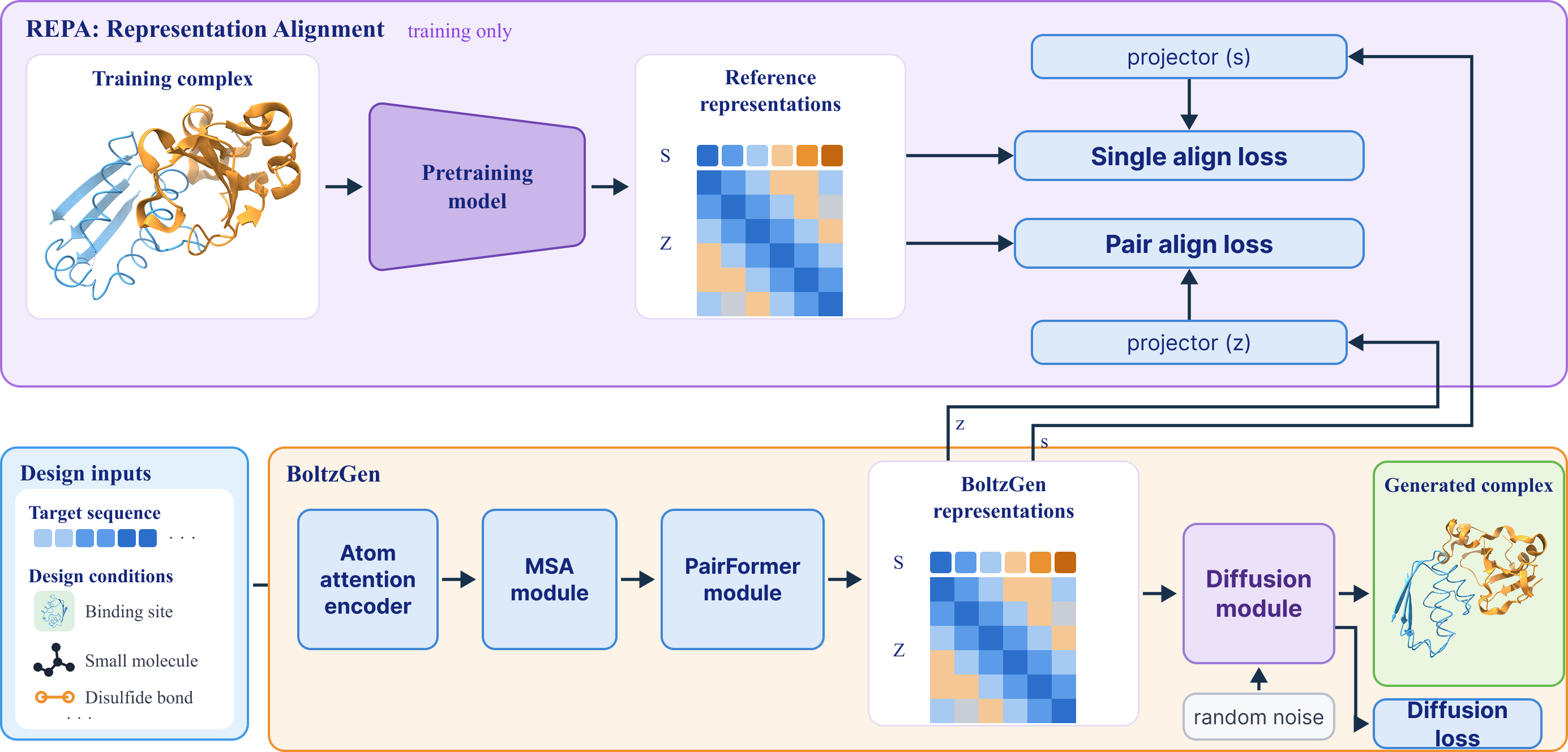}
    \caption{\textbf{Representation alignment for de novo protein binder design.}
    During training, the fixed multimodal pretraining model provides single and
    pair representation targets for known target--binder complexes. Learned
    projectors map the corresponding BoltzGen trunk representations into these
    target spaces, where the alignment loss is applied. The pretraining model and
    alignment projectors are used only during training; generation follows the
    native BoltzGen inference pipeline.}
    \label{fig:binder-alignment}
\end{figure}

We evaluate four training conditions. \textbf{BoltzGen} is the released
baseline. \textbf{+Cont.} continues training from the released checkpoint using
the same data and training schedule but without representation alignment.
\textbf{+ Ours (w/o TetSphere)} uses the same representation-alignment
procedure as the full model, but the pretraining model receives no
structure-specific TetSphere input. \textbf{Ours} uses the full pretrained
representation with TetSphere geometry. Together, these controls separate the
effects of continued training, representation alignment, and the registered
volumetric representation.

We evaluate on the ten-target ProtDBench and BoltzGen Challenge Set following
their respective standard evaluation protocols. The two benchmarks are
evaluated separately and are not pooled into a single pass rate. Within each
benchmark, target specifications, binder-length conditions, generation
budgets, random-seed policy, and screening criteria are held fixed across
methods. Reported success rates therefore measure computational screening
outcomes under the corresponding benchmark protocol rather than experimental
binding.

\paragraph{ProtDBench budgets and criteria.}
Each arm generates 32 backbones in every
$(\text{target},\text{binder-length})$ cell. The ten targets contribute 150
cells in total (BHRF1 and SC2RBD nine each; IL7RA, PDL1, TrkA, and TNFa
fifteen each; IR and H1 seventeen each; VEGFA and IL17A nineteen each), giving
4,800 backbones per arm. ProteinMPNN
(\texttt{v\_48\_020}, original weights, sampling temperature $10^{-4}$,
cysteine omitted) then designs eight sequences per backbone, yielding 38,400
sequence candidates per arm.

A candidate passes the AF2-IG Easy conjunction when normalized pLDDT $>0.8$,
interface pTM $>0.5$, unscaled interface PAE $<10.85$, and bound--unbound RMSD
$<3.5$\,\AA. The unbound prediction is run only for candidates that satisfy
the first three criteria, so the bound--unbound RMSD criterion is evaluated
conditionally. A backbone is counted as successful when at least one of its
eight designed sequences passes.

\paragraph{Challenge Set budgets and criteria.}
Each arm requests 2,000 candidates: ten targets, five generation seeds per
target, and 40 designs per seed. Each candidate consists of one backbone and
the single sequence produced by the BoltzGen inverse-folding stage, so
backbone- and sequence-level success coincide. The final hard-filter rate is
defined with respect to this fixed request budget: its numerator is the sum of
the per-target hard-filter counts, and its denominator is the 2,000 requested
candidates. We track the number of unique designed sequences separately as an
audit statistic and do not use it to renormalize the reported rate. All 2,000
requested candidates in each reported arm are sequence-unique.

The released BoltzGen hard filter is applied sequentially. A candidate must
contain no unknown residue; complex, design-subset, and binder-only backbone
RMSD must each be $\leq 2.5$\,\AA; and alanine, glycine, glutamate, leucine,
and valine fractions must not exceed $0.30$, $0.20$, $0.20$, $0.30$, and
$0.20$, respectively. Table~\ref{tab:boltzgen-design} reports this cascade as
cumulative survival: each row gives the percentage of the 2,000 requested
candidates still passing after that criterion together with all criteria above
it. The rows therefore share a single denominator, decrease monotonically, and
are comparable across arms. We report the cascade this way rather than as
per-stage retention conditional on the preceding stages, because a conditional
rate is computed over a different surviving population for each arm and cannot
distinguish a genuinely weaker stage from one that simply received more
marginal candidates from upstream.

\begin{figure}[ht]
  \centering
  \setlength{\tabcolsep}{2pt}
  \begin{tabular}{@{}ccc@{}}
    \includegraphics[width=0.323\textwidth]{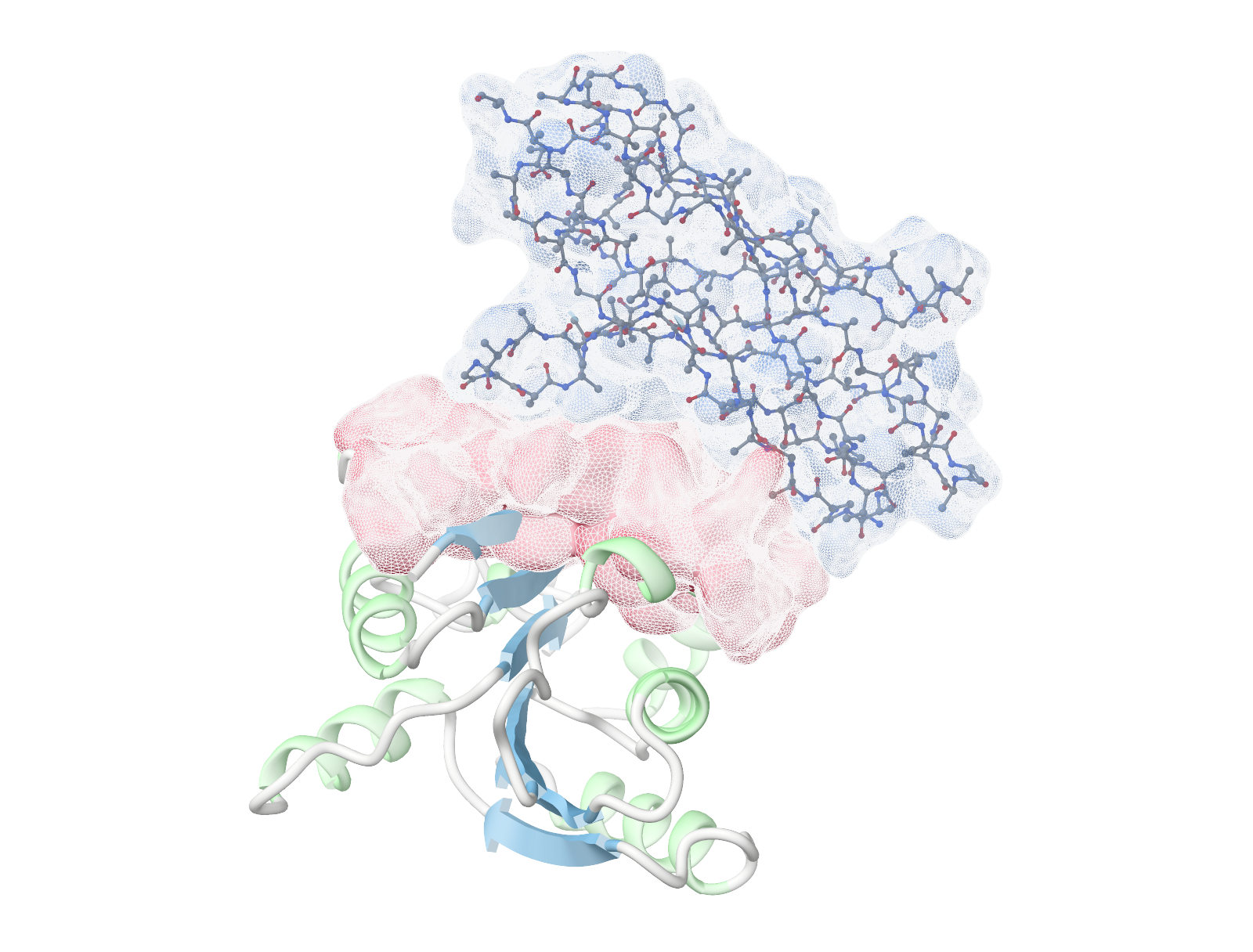} &
    \includegraphics[width=0.323\textwidth]{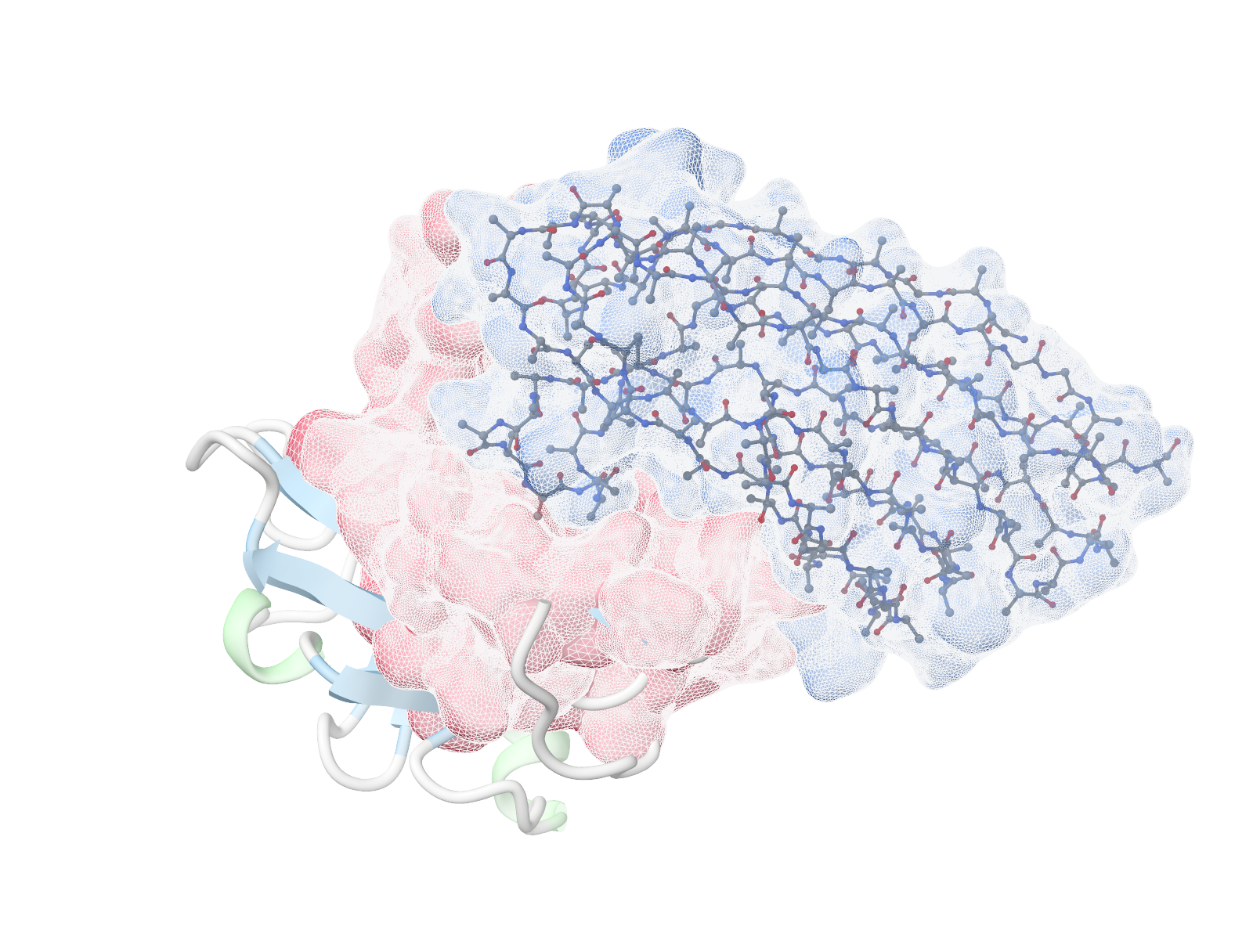} &
    \includegraphics[width=0.323\textwidth]{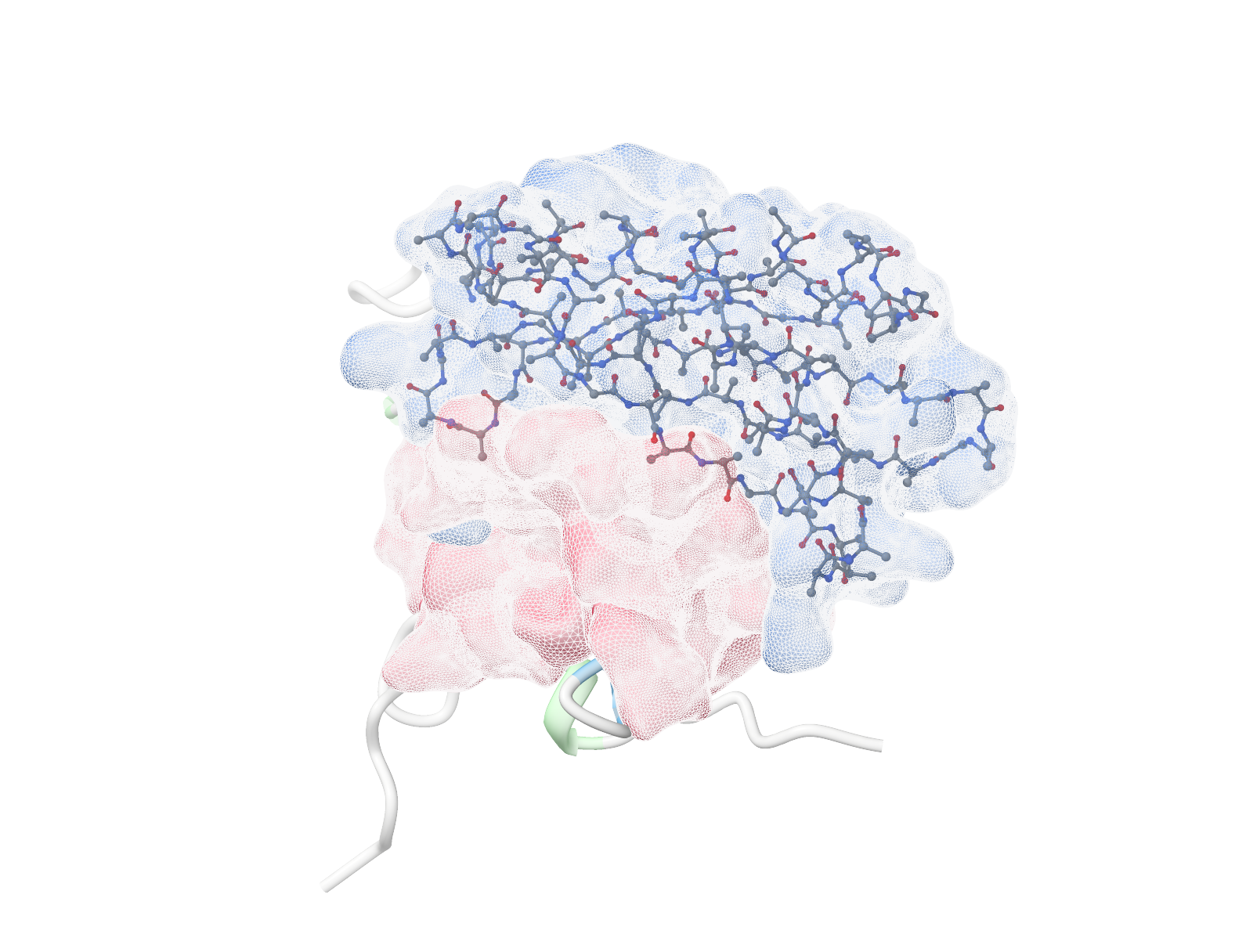} \\
    {\color[HTML]{2E7D4F}\small\bfseries (a) 1JQD} &
    {\color[HTML]{2E7D4F}\small\bfseries (b) 3QKG} &
    {\color[HTML]{2E7D4F}\small\bfseries (c) 7AAH}
  \end{tabular}
  \vspace{-6pt}
\caption{
Binders designed by our model on three BoltzGen Challenge Set targets. The designed binder is drawn as ball-and-stick inside its own TetSphere volumes (blue), and the target residues it contacts are shown as TetSphere volumes (red); the remainder of the target is a secondary-structure cartoon.
}
  \label{fig:binder-design-examples}
\end{figure}

\paragraph{Uncertainty and omitted rows.} The $\pm$ values in Table~\ref{tab:boltzgen-design} are standard deviations
from 2,000 nonparametric bootstrap replicates, with resampling performed within
targets and with ProtDBench backbones kept together with their eight designed
sequences. The target panel is held fixed. Diversity statistics are reported
without bootstrap uncertainty because resampling duplicates distort cluster
counts.

\paragraph{Structural diversity.}
Generated binder chains are extracted from each complex and clustered
separately within each target using Foldseek \texttt{easy-cluster} in TM-align
mode with a coverage threshold of $0.8$ and the stated TM-score threshold.
Cluster counts are then summed across targets.

For the diversity-adjusted cluster pass rate reported in
Table~\ref{tab:boltzgen-design}, we count Foldseek clusters among successful
designs and normalize by the total generation budget of the corresponding
benchmark: 4,800 backbones for ProtDBench and 2,000 candidates for the
Challenge Set. ProtDBench contains multiple binder-length conditions within
each target, which naturally increases its cluster count; diversity rates are
therefore intended for comparisons across training arms within a benchmark,
rather than across the two benchmarks.

For completeness, clustering all generated candidates rather than only
successful designs gives, at TM $0.6$, $17.92\%$, $22.04\%$, $36.19\%$, and
$29.56\%$ on ProtDBench and $23.95\%$, $26.30\%$, $58.15\%$, and $38.25\%$
on the Challenge Set for BoltzGen, \mbox{+Cont.},
\mbox{+Ours (w/o TetSphere)}, and Ours, respectively. At TM $0.8$, the
corresponding values are $38.81\%$, $47.92\%$, $63.90\%$, and $66.17\%$ on
ProtDBench and $42.15\%$, $44.05\%$, $75.10\%$, and $71.15\%$ on the
Challenge Set.

\end{document}